# Linking Search Space Structure, Run-Time Dynamics, and Problem Difficulty: A Step Toward Demystifying Tabu Search

**Jean-Paul Watson**                                                                     JWATSON@SANDIA.GOV
*Sandia National Laboratories*
*P.O. Box 5800, MS 1110*
*Albuquerque, NM 87185-1110 USA*

**L. Darrell Whitley**                                                                   WHITLEY@CS.COLOSTATE.EDU
**Adele E. Howe**                                                                        HOWE@CS.COLOSTATE.EDU
*Computer Science Department*
*Colorado State University*
*Fort Collins, CO 80523 USA*

## Abstract

Tabu search is one of the most effective heuristics for locating high-quality solutions to a diverse array of NP-hard combinatorial optimization problems. Despite the widespread success of tabu search, researchers have a poor understanding of many key theoretical aspects of this algorithm, including models of the high-level run-time dynamics and identification of those search space features that influence problem difficulty. We consider these questions in the context of the job-shop scheduling problem (JSP), a domain where tabu search algorithms have been shown to be remarkably effective. Previously, we demonstrated that the mean distance between random local optima and the nearest optimal solution is highly correlated with problem difficulty for a well-known tabu search algorithm for the JSP introduced by Taillard. In this paper, we discuss various shortcomings of this measure and develop a new model of problem difficulty that corrects these deficiencies. We show that Taillard's algorithm can be modeled with high fidelity as a simple variant of a straight-forward random walk. The random walk model accounts for nearly all of the variability in the cost required to locate both optimal and sub-optimal solutions to random JSPs, and provides an explanation for differences in the difficulty of random versus structured JSPs. Finally, we discuss and empirically substantiate two novel predictions regarding tabu search algorithm behavior. First, the method for constructing the initial solution is highly unlikely to impact the performance of tabu search. Second, tabu tenure should be selected to be as small as possible while simultaneously avoiding search stagnation; values larger than necessary lead to significant degradations in performance.

## 1. Introduction

Models of problem difficulty have excited considerable recent attention (Cheeseman, Kanefsky, & Taylor, 1991; Clark, Frank, Gent, MacIntyre, Tomov, & Walsh, 1996; Singer, Gent, & Smaill, 2000). These models[1] are designed to account for the variability in search cost observed for one or more algorithms on a wide range of problem instances and have yielded significant insight into the relationship between search space structure, problem difficulty, and algorithm behavior.

---

1. We refer to models of problem difficulty as *cost models* throughout this paper.

In this paper, we investigate cost models of tabu search for the job-shop scheduling problem (JSP). The JSP is an NP-hard combinatorial optimization problem and has become one of the standard and most widely studied problems in scheduling research. Tabu search algorithms are regarded as among the most effective approaches for generating high-quality solutions to the JSP (Jain & Meeran, 1999) and currently represent the state-of-the-art by a comfortable margin over the closest competition (Nowicki & Smutnicki, 2005; Błażewicz, Domschke, & Pesch, 1996). While researchers have achieved considerable advances in performance since Taillard first demonstrated the effectiveness of tabu search algorithms for the JSP in 1989, comparatively little progress has been made toward developing an understanding of how these algorithms work, i.e., characterizing the underlying high-level search dynamics, understanding why these dynamics are so effective on the JSP, and deducing how these dynamics might be modified to yield further improvements in performance.

It is well-known that the structure of the search space influences problem difficulty for tabu search and other local search algorithms (Reeves, 1998). Consequently, a dominant approach to developing cost models is to identify those features of the search space that are highly correlated with search cost. We have performed extensive analyses of the relationship between various search space features and problem difficulty for Taillard's tabu search algorithm for the JSP (Watson, Beck, Howe, & Whitley, 2001, 2003). Our findings were largely negative: many features that are widely believed to influence problem difficulty for local search are, in fact, only weakly correlated with problem difficulty. Features such as the number of optimal solutions (Clark et al., 1996), the backbone size (Slaney & Walsh, 2001), and the mean distance between random local optima (Mattfeld, Bierwirth, & Kopfer, 1999) account for less than a third of the total variability in search cost for Taillard's algorithm. In contrast, drawing from research on problem difficulty for local search and MAX-SAT (Singer et al., 2000), we found that the mean distance between random local optima and the nearest optimal solution, which we denote $\overline{d}_{lopt\text{-}opt}$, is highly correlated with problem difficulty, accounting for at least 2/3 of the total variability in search cost (Watson et al., 2003). We further demonstrated that $\overline{d}_{lopt\text{-}opt}$ accounts for much of the variability in the cost of locating sub-optimal solutions to the JSP, and for differences in the relative difficulty of "square" versus "rectangular" JSPs.

Nevertheless, the $\overline{d}_{lopt\text{-}opt}$ cost model has several shortcomings. First, the expense of computing $\overline{d}_{lopt\text{-}opt}$ limited our analyses to relatively small problem instances, raising concerns regarding scalability to more realistically sized problem instances. Second, residuals under the $\overline{d}_{lopt\text{-}opt}$ model are large for a number of problem instances, and the model is least accurate for the most difficult problem instances. Third, because the $\overline{d}_{lopt\text{-}opt}$ model provides no direct insight into the run-time behavior of Taillard's algorithm, we currently do not understand *why* $\overline{d}_{lopt\text{-}opt}$ is so highly correlated with search cost.

We introduce a novel cost model that corrects for the aforementioned deficiencies of the $\overline{d}_{lopt\text{-}opt}$ cost model. This model, which based on a detailed analysis of the run-time behavior of Taillard's algorithm, is remarkably accurate, accounting for over 95% of the variability in the cost of locating both optimal and sub-optimal solutions to a wide range of problem instances. More specifically, we establish the following results:

1. Search in Taillard's algorithm appears to be effectively restricted to a sub-space $S_{lopt+}$ of solutions that contains both local optima and solutions that are very close to (specifically, 1-2 moves away from) local optima.

2. Taillard's algorithm can be modeled with remarkable fidelity as a variant of a simple one-dimensional random walk over the $S_{lopt+}$ sub-space. The walk exhibits two notable forms of bias in the transition probabilities. First, the probability of search moving closer to (farther from) the nearest optimal solution is proportional (inversely proportional) to the current distance from the nearest optimal solution. Second, search exhibits momentum: it is more likely to move closer to (farther from) the nearest optimal solution if search previously moved closer to (farther from) the nearest optimal solution. Moreover, the random walk model accounts for at least 96% of the variability in the mean search cost across a range of test instances.

3. The random walk model is equally accurate for random, workflow, and flowshop JSPs. However, major differences exist in the number of states in the walk, i.e., the maximal possible distance between a solution and the nearest optimal solution. Differences in these maximal distances fully account for well-known differences in the difficulty of problem instances drawn from these various sub-classes.

4. The accuracy of the random walk model transfers to a tabu search algorithm based on the powerful *N5* move operator, which is more closely related to state-of-the-art tabu search algorithms for the JSP than Taillard's algorithm.

5. The random walk model correctly predicts that initiating Taillard's algorithm from high-quality starting solutions will only improve performance if those solutions are very close to the nearest optimal solution.

6. The random walk model correctly predicts that any tabu tenure larger than the minimum required to avoid search stagnation is likely to increase the fraction of the search space explored by Taillard's algorithm, and as a consequence yield a net *increase* in problem difficulty. Informally, the detrimental nature of large tabu tenures is often explained simply by observing that large tenures impact search through the loss of flexibility and the resulting inability to carefully explore the space of neighboring solutions. Our results provide a more detailed and concrete account of this phenomenon in the context of the JSP.

The remainder of this paper is organized as follows. We begin in Sections 2 and 3 with a description of the JSP, Taillard's algorithm, and the problem instances used in our analysis. The hypothesis underlying our analysis is detailed in Section 4. We summarize and critique prior research on problem difficulty for tabu search algorithms for the JSP in Section 5. Sections 6 through 9 form the core of the paper, in which we develop and validate our random walk model of Taillard's algorithm. In Section 10 we explore the applicability of the random walk model to more structured problem instances. Section 11 explores the applicability of the random walk model to a tabu search algorithm that is more representative of state-of-the-art algorithms for the JSP than Taillard's algorithm. Section 12 details two uses of the random walk model in a predictive capacity. We conclude by discussing the implications of our results and directions for future research in Section 13.

## 2. Problem and Test Instances

We consider the well-known $n \times m$ static, deterministic JSP in which $n$ jobs must be processed exactly once on each of $m$ machines (Błażewicz et al., 1996). Each job $i$ ($1 \le i \le n$) is routed through each of the $m$ machines in a pre-defined order $\pi_i$, where $\pi_i(j)$ denotes the $j$th machine ($1 \le j \le m$) in the routing order of job $i$. The processing of job $i$ on machine $\pi_i(j)$ is denoted $o_{ij}$ and is called an operation. An operation $o_{ij}$ must be processed on machine $\pi_i(j)$ for an integral duration $\tau_{ij} \ge 0$. Once initiated, processing cannot be pre-empted and concurrency on individual machines is not allowed, i.e., the machines are unit-capacity resources. For $2 \le j \le m$, $o_{ij}$ cannot begin processing until $o_{i(j-1)}$ has completed processing. The scheduling objective is to minimize the makespan $C_{max}$, i.e., the completion time of the last operation of any job. Makespan-minimization for the JSP is $NP$-hard for $m \ge 2$ and $n \ge 3$ (Garey, Johnson, & Sethi, 1976).

An instance of the $n \times m$ JSP is uniquely defined by the set of $nm$ operation durations $\tau_{ij}$ and $n$ job routing orders $\pi_i$. We define a *random* JSP as an instance generated by (1) sampling the $\tau_{ij}$ independently and uniformly from an interval $[LB, UB]$ and (2) constructing the $\pi_i$ from random permutations of the integer sequence $\zeta = 1, \ldots, m$. Most often $LB = 1$ and $UB = 99$ (Taillard, 1993; Demirkol, Mehta, & Uzsoy, 1998). The majority of JSP benchmark instances, including most found in the OR Library[2], are random JSPs.

Non-random JSPs can be constructed by imposing structure on either the $\tau_{ij}$, the $\pi_i$, or both. To date, researchers have only considered instances with structured $\pi_i$, although it is straightforward to adapt existing methods for generating non-random $\tau_{ij}$ for flow shop scheduling problems (Watson, Barbulescu, Whitley, & Howe, 2002) to the JSP. One approach to generating structured $\pi_i$ involves partitioning the set of $m$ machines into $wf$ contiguous, equally-sized subsets called *workflow partitions*. For example, when $wf = 2$, the set of $m$ machines is partitioned into two subsets containing the machines 1 through $m/2$ and $m/2 + 1$ through $m$, respectively. In such a two-partition scheme, every job must be processed on all machines in the first partition before proceeding to any machine in the second partition. No constraints are placed on the job routing orders within each partition. We refer to JSPs with $wf = 2$ simply as *workflow* JSPs. Less common are *flowshop* JSPs, where $wf = m$, i.e., all of the jobs visit the machines in the same pre-determined order.

While the presence of structure often makes scheduling problems easier to solve (Watson et al., 2002), this is not the case for JSPs with structured $\pi_i$. Given fixed $n$ and $m$, the average difficulty of problem instances – as measured by the cost required to either locate an optimal solution or to prove the optimality of a solution – is empirically proportional to $wf$. In other words, random JSPs are generally the easiest instances, while flowshop JSPs are the most difficult. Evidence for this observation stems from a wide variety of sources. For example, Storer et al. (1992) introduced sets of $50 \times 10$ random and workflow JSPs in 1992; the random JSPs were quickly solved to optimality, while the optimal makespans of all but one of the workflow JSPs are currently unknown. Similarly, the most difficult $10 \times 10$ benchmark problems, Fisher and Thompson's infamous $10 \times 10$ instance and Applegate and Cook's (1991) `orb` instances, are all "nearly" workflow or flowshop JSPs, in that the requirement that a job be processed on all machines in one workflow partition before proceeding to any machine in the next workflow partition is slightly relaxed.

---

2. http://www.brunel.ac.uk/depts/ma/research/jeb/orlib/jobshopinfo.html

Accurate cost models of local search algorithms are generally functions of the set of globally optimal solutions to a problem instance (Watson et al., 2003; Singer et al., 2000), and the cost models we develop here further emphasize this dependency. Due in part to the computational cost of enumeration, our analysis is largely restricted to sets of $6 \times 4$ and $6 \times 6$ random, workflow, and flowshop JSPs. Each set contains 1,000 instances apiece, with the $\tau_{ij}$ sampled from the interval $[1, 99]$. To assess the scalability of our cost models, we also use a set of 100 $10 \times 10$ random JSPs, where the $\tau_{ij}$ are uniformly sampled from the interval $[1, 99]$. In Section 10, we consider sets of $10 \times 10$ workflow and flowshop JSPs generated in an analogous manner. For comparative purposes with the literature, we additionally report results for various $10 \times 10$ instances found in the OR Library. Although the $10 \times 10$ OR Library instances are no longer considered particularly challenging, they have received significant historical attention and serve to validate results obtained using our own $10 \times 10$ problem set. For each instance in each of the aforementioned problem sets, a variant of Beck and Fox's (2000) constraint-directed scheduling algorithm was used to compute both the optimal makespan and the set of optimal solutions.

## 3. The Algorithm: Tabu Search and the JSP

Numerous tabu search algorithms have been developed for the JSP (Jain & Meeran, 1999). For our analysis, we select an algorithm introduced by Taillard in 1994. We implemented a variant of Taillard's algorithm, which we denote $TS_{N1}$[3], and easily reproduced results consistent with those reported by Taillard. $TS_{N1}$ is *not* a state-of-the-art tabu search algorithm for the JSP; the algorithms of Nowicki and Smutnicki (2005), Pezzella and Merelli (2000), and Barnes and Chambers (1995) yield stronger overall performance. All of these algorithms possess a core tabu search mechanism that is very similar to that found in $TS_{N1}$, but differ in the choice of move operator, the method used to generate initial solutions, and the use of long-term memory mechanisms such as reintensification.

Our choice of $TS_{N1}$ is pragmatic. Before tackling more complex, state-of-the-art algorithms, we first develop cost models of a relatively simple but representative version of tabu search and then systematically assess the influence of more complex algorithmic features on cost models of the basic algorithm. Consequently, our implementation of $TS_{N1}$ deviates from Taillard's original algorithm in three respects. First, we compute solution makespans exactly instead of using a computationally efficient estimation scheme. Second, we do not use frequency-based memory; Taillard (1994, p. 100) indicates that the benefit of such memory is largely restricted to instances requiring a very large number (i.e., > 1 million) of iterations. Third, we initiate trials of $TS_{N1}$ from random local optima (using a scheme described below) instead of those resulting from Taillard's deterministic construction method. As discussed in Section 12.1, there is strong evidence that the type of the initial solution has a negligible impact on the speed with which $TS_{N1}$ locates optimal solutions, which we take as the primary objective in our analysis.

---

3. $TS_{N1}$ is identical to the algorithm denoted $TS_{Taillard}$ in our earlier paper (Watson et al., 2003); the new notation was chosen to better convey the fact that the algorithm deviates from Taillard's original algorithm in several respects and to emphasize the relative importance of the move operator.

$TS_{N1}$ uses van Laarhoven et al.'s (1992) well-known *N1* move operator, which swaps adjacent operations on critical blocks in the schedule.[4] In our implementation, and in contrast to many local search algorithms for the JSP (Nowicki & Smutnicki, 1996), *all* pairs of adjacent critical operations are considered and not just those on a single critical path. At each iteration of $TS_{N1}$, the makespan of each neighbor of the current solution $s$ is computed and the non-tabu neighbor $s' \in N1(s)$ with the smallest makespan is selected for the next iteration; ties are broken randomly. Let $(o_{ij}, o_{kl})$ denote the pair of adjacent critical operations that are swapped in $s$ to generate $s'$, such that $o_{ij}$ appears before $o_{kl}$ in the processing order of machine $\pi_i(j)$. In the subsequent $L$ iterations, $TS_{N1}$ prevents or labels as "tabu" any move that inverts the operation pair $(o_{kl}, o_{ij})$. The idea, a variant of frequency-based memory, is to prevent recently swapped pairs of critical operations from being re-established. The scalar $L$ is known as the tabu tenure and is uniformly sampled every $1.2L_{max}$ iterations from a fixed-width interval $[L_{min}, L_{max}]$; such dynamic tabu tenures can avoid well-known cyclic search behaviors associated with fixed tabu tenures (Glover & Laguna, 1997). Let $s_{best}$ denote the best solution located during any iteration of the current run or trial of $TS_{N1}$. When $C_{max}(s') < C_{max}(s_{best})$, the tabu status of $s'$ is negated; in other words, $TS_{N1}$ employs a simple aspiration level criterion. In rare cases, the minimal neighboring makespan may be achieved by both non-tabu and tabu-but-aspired moves, in which case a non-tabu move is always accepted. If all moves in $N1(s)$ are tabu, no move is accepted for the current iteration. We observe that in the absence of tabu moves preventing improvement of the current solution's makespan, $TS_{N1}$ acts as a simple greedy descent procedure. More specifically, it is clear that the core search bias exhibited by $TS_{N1}$ is steepest-descent local search, such that there is significant pressure toward local optima.

Tabu tenure can have a major impact on performance. Based on empirical tests, Taillard defines $L_{min} = 0.8X$ and $L_{max} = 1.2X$, where $X = (n+m/2) \cdot e^{-n/5m} + N/2 \cdot e^{-5m/n}$ (Taillard, 1994); $n$ and $m$ are respectively the number of jobs and machines in the problem instance and $N = nm$. In preliminary experimentation, we observed that the resulting tenure values for our $6 \times 4$ and $6 \times 6$ problem sets (respectively $[3, 5]$ and $[4, 6]$) failed to prevent cycling or stagnation behavior. Instead, we set $[L_{min}, L_{max}]$ equal to $[6, 14]$ for all trials involving these instances and re-sample the tabu tenure every 15 iterations. For $10 \times 10$ instances we set $[L_{min}, L_{max}]$ equal to $[8, 14]$ for all trials and again re-sample the tabu tenure every 15 iterations; the specific values are taken from Taillard's research, which also ignored the aforementioned rule for trials involving $10 \times 10$ instances (Taillard, 1994). Taillard's rules are used unmodified in all trials involving larger problem instances, e.g., those analyzed below in Section 4.

### 3.1 Cost and Distance Metrics

Unlike more effective JSP move operators such as *N5* (Nowicki & Smutnicki, 1996), the *N1* operator induces search spaces that are *connected*, in that it is always possible to move from an arbitrary solution to a global optimum. Consequently, it is possible to construct a local search algorithm based on *N1* that is probabilistically approximately complete (PAC) (Hoos, 1998), such that an optimal solution will eventually be located given sufficiently large runtimes. Our experimental results suggest that $TS_{N1}$ is PAC, subject to reasonable settings for the tabu tenure; given our rules for selecting $L_{min}$ and $L_{max}$, no trial of $TS_{N1}$ failed to locate an optimal solution to any of the problem instances described in Section 2. In particular, the tabu tenure must be large enough for $TS_{N1}$ to escape local optima; using short tabu tenures, it is straightforward to construct examples where $TS_{N1}$ will become permanently trapped in the attractor basin of a single local optimum. To be provably PAC under general parameter settings, the $TS_{N1}$ algorithm would likely require modifications enabling it to accept an arbitrary move at any given iteration, allowing search to always progress toward a global optimum; Hoos (1998) discusses similar requirements in the context of PAC local search algorithms for MAX-SAT. We have not pursued such modifications because they ignore practical efficiency issues associated with poor parameter value selection, and because it is unclear how the induced randomness would impact the core tabu search dynamics.

---

4. Our notation for move operators is taken from Błażewicz (1996).

The empirical PAC property enables us to naturally define the cost required to solve a given problem instance for a *single* trial of $TS_{N1}$ as the number of iterations required to locate a globally optimal solution. In general, search cost under $TS_{N1}$ is a random variable with an approximately exponential distribution, as we discuss in Section 9. Consequently, we define the search cost for a given problem instance as either the median or mean number of iterations required to locate an optimal solution; the respective quantities are denoted by $c_{Q2}$ and $\bar{c}$. We estimate both $c_{Q2}$ and $\bar{c}$ using 1,000 independent trials. Due to the exponential nature of the underlying distribution, a large number of samples is required to achieve reasonably accurate estimates of both statistics.

Our analysis relies on the notion of the distance $D(s_1, s_2)$ between two solutions $s_1$ and $s_2$, which we take as the well-known disjunctive graph distance (Mattfeld et al., 1999). Let $\gamma(i, j, k, s)$ denote a predicate that determines whether job $j$ appears before job $k$ in the processing order of machine $i$ of solution $s$. The disjunctive graph distance $D(s_1, s_2)$ between $s_1$ and $s_2$ is then defined as

$$D(s_1, s_2) = \sum_{i=1}^{m} \sum_{j=1}^{n-1} \sum_{k=j+1}^{n} \gamma(i, j, k, s_1) \oplus \gamma(i, j, k, s_2)$$

where the symbol $\oplus$ denotes the Boolean XOR operator. Informally, the disjunctive graph distance simply captures the degree of heterogeneity observed in the machine processing sequences of two solutions. A notable property of the disjunctive graph distance is that it serves as a lower bound, which is empirically tight, on the number of *N1* moves required to transform $s_1$ into $s_2$. This is key in our analysis, as computation of the exact number of *N1* moves required to transform $s_1$ into $s_2$ is NP-hard (Vaessens, 1995). In the remainder of this paper, we use the terms "distance" and "disjunctive graph distance" synonymously.

Finally, we define a "random local optimum" as a solution resulting from the application of steepest-descent local search under the *N1* operator to a random *semi-active* solution. A semi-active solution is defined as a feasible solution (i.e., lacking cyclic ordering dependencies) in which all operations are processed at their earliest possible starting time. To construct random semi-active solutions, we use a procedure introduced by Mattfeld (1996, Section 2.2). The steepest-descent procedure employs random tie-breaking in the presence of multiple equally good alternatives and terminates once a solution $s$ is located such that $\forall s' \in N1(s)$, $C_{max}(s) \leq C_{max}(s')$.

## 4. The Run-Time Behavior of $TS_{N1}$: Motivating Observations

For a given problem instance, consider the space of feasible solutions $S$ and the sub-space $S_{lopt} \subseteq S$ containing all local optima. Due to the strong bias toward local optima induced by the core steepest-descent strategy, we expect $TS_{N1}$ to frequently sample solutions in $S_{lopt}$ during search. However, the degree to which solutions in $S_{lopt}$ are actually *representative* of solutions visited by $TS_{N1}$ is a function of both the strength of local optima attractor basins in the JSP and the specifics of the short-term memory mechanism. In particular, strong attractor basins would require $TS_{N1}$ to move far away from local optima in order to avoid search stagnation. Recently, Watson (2003) showed that the attractor basins of local optima in the JSP are surprisingly weak in general and can be escaped with high probability simply by (1) accepting a short random sequence (i.e., of length 1 or 2 elements) of monotonically worsening moves and (2) re-initiating greedy descent. In other words, relatively small perturbations are sufficient to move local search out of the attractor basin of a given local optimum in the JSP. We observe that the aforementioned procedure provides an operational definition of attractor basin strength, e.g., a specific escape probability given a "worsening" random sequence of length k, in contrast to more informal notions such as narrowness, width, or diameter.

Based on these observations, we hypothesize that search in $TS_{N1}$ is effectively restricted to the sub-space $S_{lopt+} \supset S_{lopt}$ containing both local optima and solutions that are very close to local optima in terms of disjunctive graph distance or, equivalently, the number of *N1* moves. To test this hypothesis, we monitor the *descent distance* of solutions visited by $TS_{N1}$ during search on a range of random JSPs taken from the OR Library. We define the descent distance of a candidate solution $s$ as the disjunctive graph distance $D(s, s')$ between $s$ and a local optimum $s'$ generated by applying steepest-descent under the *N1* operator to $s$. In reality, descent distance is stochastic due to the use of random tie-breaking during steepest-descent. We avoid exact characterization of descent distance for any particular $s$ and instead compute descent distance statistics over a wide range of $s$. For each problem instance, we execute $TS_{N1}$ for one million iterations, computing the descent distance of the current solution at each iteration and recording the resulting time-series; trials of $TS_{N1}$ are terminated once an optimal solution is encountered and re-started from a random local optimum. Several researchers have introduced measures that are conceptually related to descent distance, but are in contrast based on differences in solution fitness. For example, search *depth* (Hajek, 1988) is defined as the minimal increase in fitness (assuming minimization) that must be accepted in order to escape a local optimum attractor basin. Similarly, Schuurmans and Southey (2001) define search depth as the difference between the fitness of a solution $s$ and that of a global optimum $s^*$.

Summary statistics for the resulting descent distances are reported in Table 1; for comparative purposes, we additionally report the mean descent distance observed for one million random semi-active solutions. The mean and median descent distance statistics indicate that $TS_{N1}$ consistently remains only 1–2 moves away from local optima, *independent* of problem size. Although search is occasionally driven very far from local optima, such events are rare - as corroborated by the low standard deviations. Empirical evidence suggests that large-distance events are not due to the existence of local optima with very deep attractor basins, but are rather an artifact of $TS_{N1}$'s short-term memory mechanism. These results support our hypothesis that search in $TS_{N1}$ is largely restricted to the sub-space of feasible solutions containing both local optima and solutions that are very close (in terms of disjunctive graph distance) to local optima. Two factors enable this behavior: the strong bias toward local optima that is induced by the core steepest-descent strategy of $TS_{N1}$ and the relative weakness of attractor basins in the JSP. These factors also enable $TS_{N1}$ to ignore substantial proportions of the search space, as the mean descent distance under $TS_{N1}$ is only a fraction of the descent distance of random semi-active solutions.

| Size | Instance | Descent Distance Under $TS_{N1}$ | | | | Mean Descent Distance for Random Solutions |
|---|---|---|---|---|---|---|
| | | Median | Mean | Std. Dev. | Max. | |
| $10 \times 10$ | **la16** | 1 | 1.84 | 1.78 | 20 | 12.43 |
| $10 \times 10$ | **la17** | 1 | 1.96 | 1.86 | 16 | 13.03 |
| $10 \times 10$ | **la18** | 2 | 2.06 | 1.94 | 18 | 13.42 |
| $10 \times 10$ | **la19** | 2 | 2.15 | 1.94 | 18 | 13.99 |
| $10 \times 10$ | **la20** | 2 | 2.22 | 1.94 | 20 | 13.00 |
| $10 \times 10$ | **abz5** | 2 | 2.16 | 1.95 | 16 | 13.96 |
| $10 \times 10$ | **abz6** | 2 | 2.12 | 1.89 | 16 | 12.95 |
| $15 \times 15$ | **ta01** | 1 | 1.95 | 1.97 | 18 | 24.38 |
| $20 \times 15$ | **ta11** | 1 | 1.51 | 1.83 | 21 | 30.07 |
| $20 \times 20$ | **ta21** | 2 | 2.28 | 2.24 | 24 | 34.10 |
| $30 \times 15$ | **ta31** | 1 | 2.00 | 2.20 | 20 | 39.27 |
| $30 \times 20$ | **ta41** | 1 | 1.68 | 2.02 | 36 | 46.19 |
| $50 \times 15$ | **ta51** | 1 | 1.86 | 2.35 | 28 | 45.11 |
| $50 \times 20$ | **ta61** | 2 | 2.25 | 2.56 | 60 | 55.50 |
| $100 \times 20$ | **ta71** | 1 | 2.50 | 3.16 | 40 | 71.26 |

Table 1: Descent distance statistics for $TS_{N1}$ on select random JSPs from the OR Library. Statistics are taken over time-series of length one million iterations. Results for random solutions are computed using one million random semi-active solutions.

The primary purpose of short-term memory in tabu search is to enable escape from local optima (Glover & Laguna, 1997). We recall that $TS_{N1}$ does not employ any long-term memory mechanism. Given (1) the absence of an explicit high-level search strategy and (2) the lack of any *a priori* evidence to suggest that $TS_{N1}$ is biased toward specific, e.g., high-quality, regions of $S_{lopt+}$, we propose the following hypothesis: $TS_{N1}$ *is simply performing a random walk over the $S_{lopt+}$ sub-space*. If true, problem difficulty should be correlated with $|S_{lopt+}|$, the size of the $S_{lopt+}$ sub-space. Implicit in this assertion is the assumption that the connectivity in $S_{lopt+}$ is sufficiently regular such that optimal solutions are not isolated relative to the rest of the search space, e.g., only reachable through a few highly improbable pathways. Finally, we observe that the presence of multiple optimal solutions reduces the proportion of $S_{lopt+}$ that must be explored, on average, before a globally optimal solution is encountered. Consequently, we focus instead on the *effective* size of $S_{lopt+}$, which we denote by $|S_{lopt+}|'$. For now, this measure is only an abstraction used to capture the notion that problem difficulty is a function of $|S_{lopt+}|$ and the number and distribution of globally optimal solutions within that sub-space. Specific estimates for $|S_{lopt+}|'$ are considered below in Sections 5 through 7.

## 5. Static Cost Models: A Summary and Critique of Prior Research

With the exception of research on phase transitions in mean instance difficulty (Clark et al., 1996), all existing cost models of local search algorithms are based on structural analyses of the underlying fitness landscapes. Informally, a fitness landscape is a vertex-weighted graph in which the verticies represent candidate solutions, the vertex weights represent the fitness or worth of solutions, and the edges capture solution adjacency relationships induced by a neighborhood or move operator (c.f., Reeves, 1998; Stadler, 2002).

Previously, we analyzed the relationship between various fitness landscape features and problem difficulty for $TS_{N1}$ (Watson et al., 2001, 2003). We used regression methods to construct statistical models relating one or more (optionally transformed) landscape features, e.g., the logarithm of the number of optimal solutions, to the transformed cost $log_{10}(c_{Q2})$ required to locate optimal solutions to $6 \times 4$ and $6 \times 6$ random JSPs. Because they are based on static, time-invariant features of the fitness landscape, we refer to these models as *static* cost models. The accuracy of a static cost model can be quantified as the regression $r^2$, i.e., the proportion of the total variability in search cost accounted for by the model. Most of the models we considered were based on simple linear regression over unitary features, such that the $r^2$ captured variability under the assumption of a linear functional relationship between landscape features and search cost. We found that the accuracy of static cost models based on well-known landscape features such as the number of optimal solutions (Clark et al., 1996), the backbone size (Slaney & Walsh, 2001), and the average distance between random local optima (Mattfeld et al., 1999) was only weak-to-moderate, with $r^2$ ranging from 0.22 to 0.54. Although not reported, we additionally investigated measures such as fitness-distance correlation (Boese, Kahng, & Muddu, 1994) and landscape correlation length (Stadler, 2002) – measures that are much more commonly investigated in operations research and evolutionary computing than in AI – and obtained even lower $r^2$ values.

Drawing from research on MAX-SAT (Singer et al., 2000), we then demonstrated that a static cost model based on the mean distance between random local optima and the nearest optimal solution, which we denote $\overline{d}_{lopt\text{-}opt}$, is significantly more accurate, yielding $r^2$ values of 0.83 and 0.65 for $6 \times 4$ and $6 \times 6$ random JSPs, respectively. As shown in the left side of Figure 1, the actual $c_{Q2}$ for $6 \times 6$ random JSPs is typically within a factor of 10 (i.e., no more than 10 times and no less than 1/10) of the predicted $c_{Q2}$, although in a few exceptional cases the observed differences exceed a factor of 100. Additionally, we showed that the $\overline{d}_{lopt\text{-}opt}$ model accounts for most of the variability in the cost required to locate *sub-optimal* solutions to these same problem instances and provides an explanation for differences in the relative difficulty of "square" ($n/m \approx 1$) versus "rectangular" ($n/m \gg 1$) random JSPs.

For the variant of the $\overline{d}_{lopt\text{-}opt}$ measure associated with MAX-SAT, Singer (2000, p. 67) speculates that instances with large $\overline{d}_{lopt\text{-}opt}$ are more difficult due to "... initial large distance from the [optimal] solutions or the extensiveness of the ... area in which the [optimal] solutions lie, or a combination of these factors." Building on this observation, we view $\overline{d}_{lopt\text{-}opt}$ as a concrete measure of $|S_{lopt+}|'$, specifically because $\overline{d}_{lopt\text{-}opt}$ simultaneously accounts for *both* the size of $S_{lopt+}$ and the distribution of optimal solutions within $S_{lopt+}$.

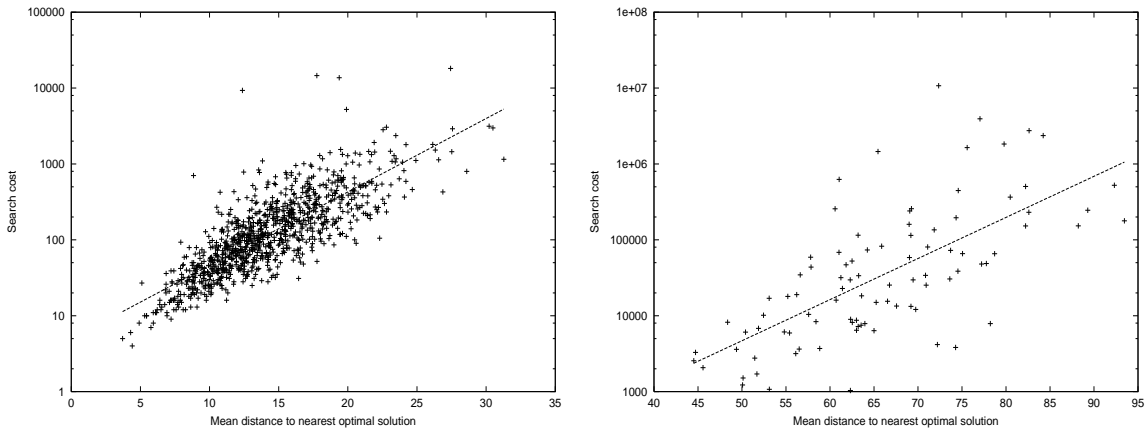

Figure 1: Scatter-plots of $\overline{d}_{lopt\text{-}opt}$ versus $c_{Q2}$ for $6 \times 6$ (left figure) and $10 \times 10$ (right figure) random JSPs; the least-squares fit lines are super-imposed.

In doing so, we assume that random local optima are representative of solutions in $S_{lopt+}$, which is intuitively justifiable given the low mean search depth observed under $TS_{N1}$.[5]

In support of our view concerning the key role of $|S_{lopt+}|'$ in problem difficulty, we observe that the other static cost models of $TS_{N1}$ considered by Watson and his colleagues (2001, 2003) are based on landscape features (the backbone size, the number of optimal solutions, or the average distance between local optima) that quantify either the size of $S_{lopt+}$ or the number/distribution of optimal solutions, but not both. In other words, the underlying measures fail to capture one of the two key dimensions of $|S_{lopt+}|'$.

Despite its explanatory power, we previously identified several deficiencies of the $\overline{d}_{lopt\text{-}opt}$ cost model (Watson et al., 2003). First, the model is least accurate for the most difficult problem instances within a fixed-size group. Second, the model fails to account for a non-trivial proportion ($\approx 1/3$) of the variability in problem difficulty for $6 \times 6$ random JSPs. Third, model accuracy fails to transfer to more structured workflow JSPs.

### 5.1 An Analysis of Scalability

Differences in the accuracy of the $\overline{d}_{lopt\text{-}opt}$ model on $6 \times 4$ and $6 \times 6$ random JSPs also raise concerns regarding scalability to larger, more realistically sized problem instances. Empirically, we have observed that the mean number of optimal solutions in random JSPs grows rapidly with increases in problem size. When coupled with the difficulty of "square" instances with $n \approx m > 10$, the resulting cost of computing both $\overline{d}_{lopt\text{-}opt}$ and $c_{Q2}$ previously restricted our analysis to $6 \times 4$ and $6 \times 6$ random JSPs. However, with newer microprocessors, we are now able to assess the accuracy of the $\overline{d}_{lopt\text{-}opt}$ cost model on larger random JSPs.

We compute $\overline{d}_{lopt\text{-}opt}$ for the 92 of our 100 $10 \times 10$ random JSPs with $\leq 50$ million optimal solutions; the computation is unpractical for the remaining 8 instances. Estimates of $\overline{d}_{lopt\text{-}opt}$ are based on 5,000 random local optima. We show a scatter-plot of $\overline{d}_{lopt\text{-}opt}$ versus $c_{Q2}$ for these problem instances in the right side of Figure 1. The $r^2$ value of the corresponding regression model is 0.46, which represents a 33% decrease in model accuracy relative to the 6 × 6 problem set. This result demonstrates the failure of the $\overline{d}_{lopt\text{-}opt}$ model to scale to larger JSPs. We observe similar drops in accuracy for static cost models based on the number of optimal solutions, the backbone size, and the mean distance between random local optima (Watson, 2003). Unfortunately, we cannot currently assess larger rectangular instances due to the vast numbers (i.e., tens of billions) of optimal solutions.

---

5. As discussed in Section 6, empirical data obtained during our search for more accurate cost models of $TS_{N1}$ ultimately forces us to retract, or more precisely modify, this assumption. However, restrictions on the $S_{lopt+}$ sub-space still play a central role in all subsequent cost models.

## 6. Accounting for Search Bias: A Quasi-Dynamic Cost Model

The deficiencies of the $\overline{d}_{lopt\text{-}opt}$ cost model indicate that either (1) $\overline{d}_{lopt\text{-}opt}$ is not an entirely accurate measure of $|S_{lopt+}|'$ or (2) our random walk hypothesis is incorrect, i.e., $|S_{lopt+}|'$ is not completely indicative of problem difficulty. We now focus on the first alternative, with the goal of developing a more accurate measure of $|S_{lopt+}|'$ than $\overline{d}_{lopt\text{-}opt}$. Instead of random local optima, we instead consider the set of solutions visited by $TS_{N1}$ *during* search. We refer to the resulting cost model as a *quasi-dynamic* cost model. The "quasi-dynamic" modifier derives from the fact that although algorithm dynamics are taken into account, an explicit model of run-time behavior is not constructed.

We develop our quasi-dynamic cost model of $TS_{N1}$ by analyzing the distances between solutions visited during search and the corresponding nearest optimal solutions. Let $d_{opt}(s)$ denote the distance between a solution $s$ and the nearest optimal solution, i.e., $d_{opt}(s) = min_{x \in S^*} D(x, s)$ where $S^*$ denotes the set of optimal solutions. Let $X_{tabu}$ denote the set of solutions visited by $TS_{N1}$ during an extended run on a given problem instance, and let $X_{rlopt}$ denote a set of random local optima. We then define $\overline{d}_{tabu\text{-}opt}$ ($\overline{d}_{lopt\text{-}opt}$) as the mean distance $d_{opt}(s)$ between solutions $s \in X_{tabu}$ ($s \in X_{rlopt}$) and the nearest optimal solution.

Figure 2 shows empirical distributions of $d_{opt}(s)$ for the $X_{rlopt}$ and $X_{tabu}$ of two 10 × 10 random JSPs. Both types of distribution are generally symmetric and Gaussian-like, although we infrequently observe skewed distributions both with and without heavier-than-Gaussian tails. Deviations from the Gaussian ideal are more prevalent in the smaller 6 × 4 and 6×6 problem sets. In all of our test instances, $\overline{d}_{tabu\text{-}opt} < \overline{d}_{lopt\text{-}opt}$, i.e., $TS_{N1}$ consistently visits solutions that on average are closer to an optimal solution than randomly generated local optima. Similar observations hold for solution *quality*, such that solutions in $X_{tabu}$ consistently possess lower makespans than solutions in $X_{rlopt}$.

The histograms shown in Figure 2 serve as illustrative examples of two types of search bias exhibited by $TS_{N1}$. First, search is strongly biased toward solutions that are an "average" distance between the nearest optimal solution and solutions that are maximally distant from the nearest optimal solution. Second, random local optima are *not* necessarily representative of the set of solutions visited during search, contradicting the assumption we stated previously in Section 5. Although search in $TS_{N1}$ *is* largely restricted to $S_{lopt+}$, there potentially exist large portions of $S_{lopt+}$ – for reasons we currently do not fully understand – that $TS_{N1}$ is unlikely visit. Failure to account for these unexplored regions will necessarily yield conservative estimates of $|S_{lopt+}|'$. We observe that these results do not contradict our random walk hypothesis; rather, we still assert that $TS_{N1}$ is performing a random walk over a potentially restricted sub-set of $S_{lopt+}$.

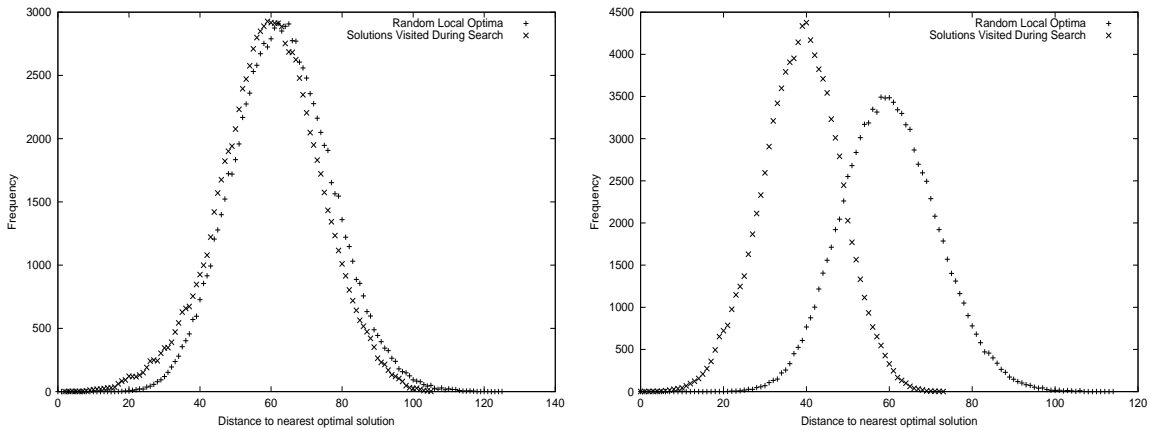

Figure 2: Histograms of the distance to the nearest optimal solution for both random local optima and solutions visited by $TS_{N1}$ for two example $10 \times 10$ random JSPs (each figure corresponds to a unique problem instance).

We believe that the deficiencies of the $\overline{d}_{lopt\text{-}opt}$ model are due in large part to the failure of the underlying measure to accurately depict the sub-space of solutions likely to be explored by $TS_{N1}$. In contrast, the $\overline{d}_{tabu\text{-}opt}$ measure by definition accounts for the set of solutions likely to be visited by $TS_{N1}$. Consequently, we hypothesize that a quasi-dynamic cost model based on the $\overline{d}_{tabu\text{-}opt}$ measure should yield significant improvements in accuracy over the static $\overline{d}_{lopt\text{-}opt}$ cost model. As evidence for this hypothesis, we observe that although discrepancies between the distributions of $d_{opt}(s)$ for random local optima and solutions visited by $TS_{N1}$ were minimal in our $6 \times 4$ problem sets, significant differences were observed in the larger $6 \times 6$ and $10 \times 10$ problem sets – the same instances for which the $\overline{d}_{lopt\text{-}opt}$ model is least accurate. To further illustrate the magnitude of the differences, we observe that for the 42 of our $10 \times 10$ random JSPs with $\leq 100{,}000$ optimal solutions, $\overline{d}_{tabu\text{-}opt}$ is on average 37% lower than $\overline{d}_{lopt\text{-}opt}$. For the same instances, the solutions in $X_{tabu}$ on average possess a makespan 13% lower than those of solutions in $X_{rlopt}$.

We now quantify the accuracy of the $\overline{d}_{tabu\text{-}opt}$ quasi-dynamic cost model on $6 \times 4$, $6 \times 6$, and $10 \times 10$ random JSPs. For any given instance, we construct $X_{tabu}$ using solutions visited by $TS_{N1}$ over a variable number of independent trials. A trial is initiated from a random local optimum and terminated once a globally optimal solution is located. The termination criterion is imposed because there exist globally optimal solutions from which no moves are possible under the N1 move operator (Nowicki & Smutnicki, 1996). We terminate the entire process, including the current trial, once $|X_{tabu}|=100{,}000$. The resulting $X_{tabu}$ are then used to compute $\overline{d}_{tabu\text{-}opt}$; the large number of samples is required to achieve reasonably accurate estimates of this statistic.

Scatter-plots of $\overline{d}_{tabu\text{-}opt}$ versus $c_{Q2}$ for the $6 \times 4$ and $6 \times 6$ problem sets are respectively shown in the upper left and upper right sides of Figure 3. Regression models of $\overline{d}_{tabu\text{-}opt}$ versus $log_{10}(c_{Q2})$ yield respective $r^2$ values of 0.84 and 0.78, corresponding to 4% and 20% increases in accuracy relative to the $\overline{d}_{lopt\text{-}opt}$ cost model. The actual $c_{Q2}$ typically deviate from the predicted $c_{Q2}$ by no more than a factor of five and we observe fewer and less extreme large-residual instances than under the $\overline{d}_{lopt\text{-}opt}$ model (only three out of 2,000 data points differ by more than a factor of 10.)

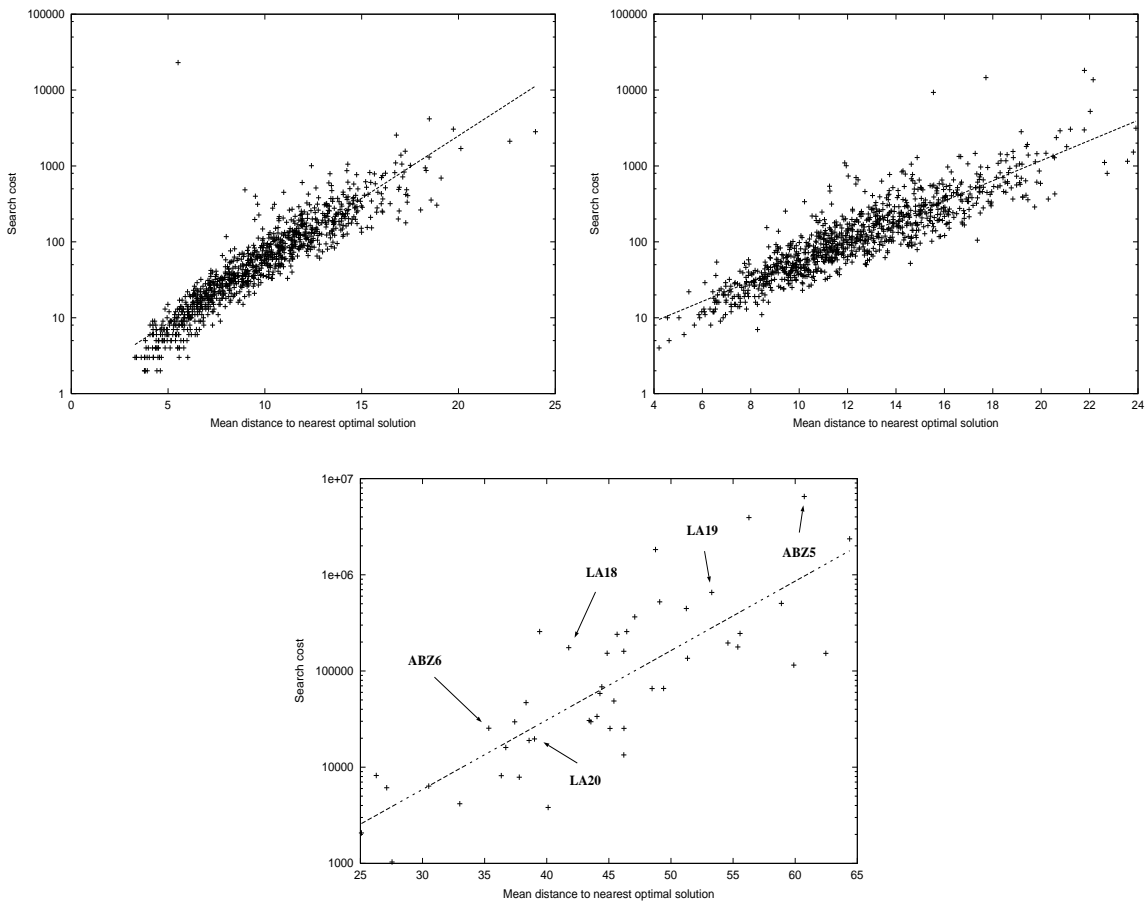

Figure 3: Scatter-plots of $\overline{d}_{tabu\text{-}opt}$ versus search cost $c_{Q2}$ for $6 \times 4$ (upper left figure), $6 \times 6$ (upper right figure), and $10 \times 10$ (lower figure) random JSPs; the least-squares fit lines are super-imposed.

For the set of 42 $10 \times 10$ random JSPs with $\leq 100,000$ optimal solutions,[6] a regression model of $\overline{d}_{tabu\text{-}opt}$ versus $log_{10}(c_{Q2})$ yields an $r^2$ value of 0.66 (see the lower portion of Figure 3); the computation of $\overline{d}_{tabu\text{-}opt}$ is unpractical for the remaining instances. The resulting $r^2$ is 41% greater than that observed for the $\overline{d}_{lopt\text{-}opt}$ model on the same instances. Further, the actual $c_{Q2}$ is typically within a factor of 5 of the predicted $c_{Q2}$ and in no case is the discrepancy larger than a factor of 10. We have also annotated the scatter-plot shown in the lower portion of Figure 3 with data for those five of the seven $10 \times 10$ random JSPs present in the OR Library with $\leq 100,000$ optimal solutions. The `abz5` and `la19` instances have been found to be the most difficult in this set (Jain & Meeran, 1999), which is consistent with the observed values of $\overline{d}_{tabu\text{-}opt}$.

---

6. Our selection criterion does *not* lead to a clean distinction between easy and hard problem instances; the hardest $10 \times 10$ instance has approximately 1.5 million optimal solutions. However, instances with $\leq 100,000$ optimal solutions are generally more difficult, with a median $c_{Q2}$ of 65,710, versus 13,291 for instances with more than 100,000 optimal solutions.

In conclusion, $TS_{N1}$ is highly unlikely to visit large regions of the search space for many problem instances. As a consequence, measures of $|S_{lopt+}|'$ based on purely random local optima are likely to be conservative and inaccurate, providing a partial explanation for the failures of the $\overline{d}_{lopt\text{-}opt}$ model. In contrast, the $\overline{d}_{tabu\text{-}opt}$ measure by definition accounts for this phenomenon, yielding a more accurate measure of $|S_{lopt+}|'$ and a more accurate cost model. However, despite the significant improvements in accuracy, the $\overline{d}_{lopt\text{-}opt}$ and $\overline{d}_{tabu\text{-}opt}$ do share two fundamental deficiencies: accuracy still fails to scale to larger problem instances, and the models provide no direct insight into the relationship between the underlying measures and algorithmic run-time dynamics.

## 7. A Dynamic Cost Model

The $\overline{d}_{lopt\text{-}opt}$ and $\overline{d}_{tabu\text{-}opt}$ cost models provide strong evidence that $TS_{N1}$ is effectively performing a random walk over a potentially restricted subset of the $S_{lopt+}$ sub-space. However, we have yet to propose any specific details, e.g., the set of states in the model or the qualitative nature of the transition probabilities. The dynamic behavior of any *memoryless* local search algorithm, e.g., iterated local search (Lourenco, Martin, & Stützle, 2003) and simulated annealing (Kirkpatrick, Gelatt, & Vecchi, 1983), can, at least in principle, be modeled using Markov chains: the set of feasible solutions is known, the transition probabilities between neighboring solutions can be computed, and the Markov property is preserved. Local search algorithms augmented with memory, e.g., tabu search, can also be modeled as Markov chains by embedding the contents of memory into the state definition, such that the Markov property is preserved. Although exact, the resulting models generally require at least (depending on the complexity of the memory) an exponential number of states – $\mathcal{O}(2^{m \cdot \binom{n}{2}})$ in the JSP – and therefore provide little insight into the qualitative nature of the search dynamics. The challenge is to develop aggregate models in which large numbers of states are grouped into meta-states, yielding more tractable and consequently understandable Markov chains.

### 7.1 Definition

To model the impact of short-term memory on the behavior of $TS_{N1}$, we first analyze how search progresses either toward or away from the nearest optimal solution. In Figure 4, we show a time-series of the distance to the nearest optimal solution for both a random walk under the *N1* move operator and $TS_{N1}$ on a typical $10 \times 10$ random JSP. We obtain similar results on a sampling of random $6 \times 4$ and $6 \times 6$ JSPs, in addition to a number of structured problem instances. The random walk exhibits minimal short-term trending behavior, with search moving away from or closer to an optimal solution with roughly equal probability. In contrast, we observe strong regularities in the behavior of $TS_{N1}$. The time-series shown in the right side of Figure 4 demonstrates that $TS_{N1}$ is able to maintain search gradients for extended periods of time. This observation leads to the following hypothesis: the short-term memory mechanism of $TS_{N1}$ acts to consistently bias search either toward or away from the nearest optimal solution.

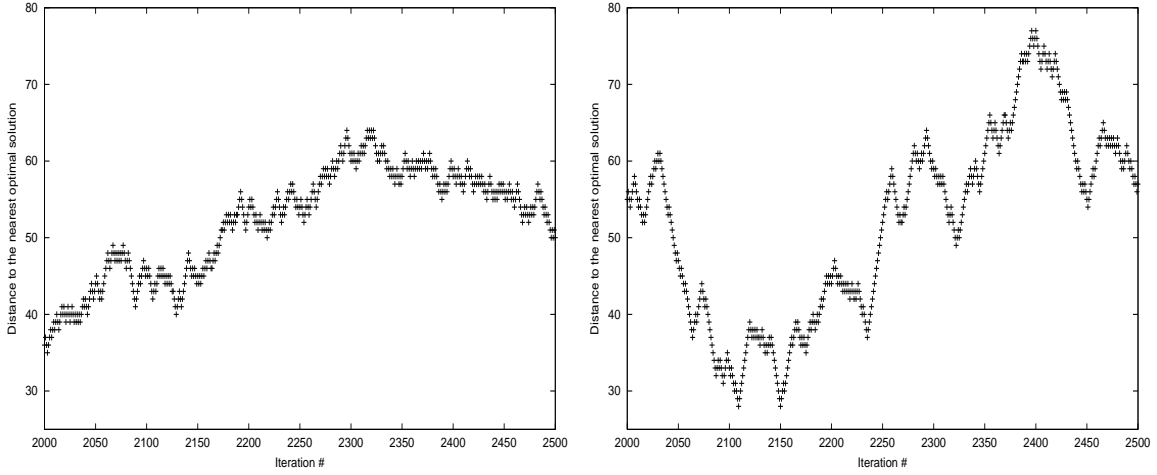

Figure 4: Time-series of the distance to the nearest optimal solution for solutions visited by a random walk (left figure) and $TS_{N1}$ (right figure) for a $10 \times 10$ random JSP.

Based on this hypothesis, we define a state $S_{x,i}$ in our Markov model of $TS_{N1}$ as a pair representing both (1) the set of solutions distance $i$ from the nearest optimal solution and (2) the current search gradient $x$. We denote the numeric values $x \in [-1, 0, 1]$ with the symbols *closer*, *equal*, and *farther*, respectively. In effect, we are modeling the impact of short-term memory as a simple scalar and embedding this scalar into the state definition. Next, we denote the maximum possible distance from an arbitrary solution to the nearest optimal solution by $D_{max}$. Finally, let the conditional probability $P(S_{j,x'}|S_{i,x})$ denote the probability of simultaneously altering the search gradient from $x$ to $x'$ and moving from a solution distance $i$ from the nearest optimal solution to a solution distance $j$ from the nearest optimal solution. The majority of these probabilities equal 0, specifically for any pair of states $S_{j,x'}$ and $S_{i,x}$ where $|i - j| > 1$ or when simultaneous changes in both the gradient and the distance to the nearest optimal solution are logically impossible, e.g., from state $S_{i,closer}$ to state $S_{i+1,closer}$. For each $i$, $1 \leq i \leq D_{max}$, there exist at most the following nine non-zero transition probabilities:

- $P(S_{i-1,closer}|S_{i,closer})$, $P(S_{i,equal}|S_{i,closer})$, and $P(S_{i+1,farther}|S_{i,closer})$
- $P(S_{i-1,closer}|S_{i,equal})$, $P(S_{i,equal}|S_{i,equal})$, and $P(S_{i+1,farther}|S_{i,equal})$
- $P(S_{i-1,closer}|S_{i,farther})$, $P(S_{i,equal}|S_{i,farther})$, and $P(S_{i+1,farther}|S_{i,farther})$

The probabilities $P(S_{j,x'}|S_{i,x})$ are also subject to the following total-probability constraints:

- $P(S_{i-1,closer}|S_{i,closer}) + P(S_{i,equal}|S_{i,closer}) + P(S_{i+1,farther}|S_{i,closer}) = 1$
- $P(S_{i-1,closer}|S_{i,equal}) + P(S_{i,equal}|S_{i,equal}) + P(S_{i+1,farther}|S_{i,equal}) = 1$
- $P(S_{i-1,closer}|S_{i,farther}) + P(S_{i,equal}|S_{i,farther}) + P(S_{i+1,farther}|S_{i,farther}) = 1$

To complete the Markov model of $TS_{N1}$, we create a reflecting barrier at $i = D_{max}$ and an absorbing state at $i = 0$ by respectively imposing the constraints $P(S_{0,closer}|S_{0,closer}) = 1$ and $P(S_{D_{max}-1,closer}|S_{D_{max},farther}) + P(S_{D_{max},equal}|S_{D_{max},farther}) = 1$. These constraints yield three isolated states: $S_{0,equal}$, $S_{0,farther}$, and $S_{D_{max},closer}$. Consequently the Markov model consists of exactly $3 \cdot D_{max}$ states.

We conclude by noting that an aggregated random walk model of $TS_{N1}$ (or any other local search algorithm) will *not* capture the full detail of the underlying search process. In particular, the partition induced by aggregating JSP solutions based on their distance to the nearest optimal solution is not *lumpable* (Kemeny & Snell, 1960); distinct solutions at identical distances to the nearest optimal solution have different transition probabilities for moving closer to and farther from the nearest optimal solution, due to both (1) unique numbers and distributions of infeasible neighbors and (2) unique distributions of neighbor makespans. Thus, the question we are posing is whether there exist sufficient regularities in the transition probabilities for solutions within a given partition such that it is possible to closely approximate the behavior of the full Markov chain using a reduced-order chain.

### 7.2 Parameter Estimation

We estimate the Markov model parameters $D_{max}$ and the set of $P(S_{j,x'}|S_{i,x})$ by sampling a subset of solutions visited by $TS_{N1}$. For a given problem instance, we obtain at least $S_{min}$ and at most $S_{max}$ *distinct* solutions at each distance $i$ from the nearest optimal solution, where $2 \leq i \leq rint(\overline{d}_{lopt\text{-}opt})$.[7] For the $6 \times 4$ and $6 \times 6$ problem sets, we let $S_{min} = 50$ and $S_{max} = 250$; for the $10 \times 10$ set, we let $S_{min} = 50$ and $S_{max} = 500$. These values of $S_{min}$ and $S_{max}$ are large enough to ensure that artificially isolated states are not generated due to an insufficient number of samples. Individual trials of $TS_{N1}$ are executed until a globally optimal solution is located, at which point a new trial is initiated. The process repeats until at least $S_{min}$ samples are obtained for each distance $i$ from the nearest optimal solution, $2 \leq i \leq rint(\overline{d}_{lopt\text{-}opt})$, at which point the current algorithmic trial is immediately terminated.

The upper bound $S_{max}$ is imposed to mitigate the impact of solutions that are statistically unlikely to be visited by $TS_{N1}$ during any *individual* trial, but are nonetheless encountered with non-negligible probability when executing the large number of trials that are required to achieve the sampling termination criterion. Informally, $S_{max}$ allows us to ensure that only truly representative solutions are included in the sample set. Candidate solutions are only considered for inclusion every 100 iterations for the smaller $6 \times 4$ and $6 \times 6$ problem sets, and every 200 iterations for the larger $10 \times 10$ problem set. Such periodic sampling ensures that the collected samples are uncorrelated; the specific sampling intervals are based on estimates of the landscape correlation length (Mattfeld et al., 1999), i.e., the expected number of iterations of a random walk after which solution fitness is uncorrelated. Candidate solutions are accepted in order of appearance, i.e., the first $S_{min}$ distinct solutions encountered at a given distance $i$ are always retained, and are discarded once the number of prior samples at distance $i$ exceeds $S_{max}$. For each sampled solution at distance $i$ from the nearest optimal solution and search gradient $x$, we track the distance $j$ and gradient $x'$ for the solution in the subsequent iteration.

---

7. The function $rint(x)$ is defined as $rint(x) = \lfloor x + 0.5 \rfloor$, which rounds to the nearest integer.

Let $\#(S_{i,x})$ and $\#(S_{j,x'}|S_{i,x})$ respectively denote the total number of observed samples in state $S_{i,x}$ and the total number of observed transitions from a state $S_{i,x}$ to a state $S_{j,x'}$. Estimates of the transition probabilities are computed using the obvious formulas, e.g., $P(S_{i-1,closer}|S_{i,closer}) = \#(S_{i-1,closer}|S_{i,closer})/\#(S_{i,closer})$. We frequently observe at least $S_{min}$ samples for distances $i > rint(\overline{d}_{lopt\text{-}opt})$. To estimate $D_{max}$, we first determine the minimal $X$ such that the number of samples at distance $X$ is less than $S_{min}$, i.e., the smallest distance at which samples are not consistently observed. We then define $D_{max} = X - 1$; omission of states $S_{i,x}$ with $i > X$ has negligible impact on model accuracy. Finally, we observe that our estimates of both the $P(S_{j,x'}|S_{i,x})$ and $D_{max}$ are largely insensitive to both the initial solution and the sequence of solutions visited during the various trials, i.e., the statistics appear to be isotropic.

The aforementioned process is online in that the computed parameter estimates are based on solutions actually visited by $TS_{N1}$. Ideally, parameter estimates could be derived independently of the algorithm under consideration, for example via an analysis of random local optima. However, two factors conspire to prevent such an approach in the JSP. First, as shown in Section 6, random local optima are typically not representative of solutions visited by $TS_{N1}$ during search, and we currently do not fully understand the root cause of this phenomenon (although preliminary evidence indicates it is due in large part to the distribution of infeasible solutions within the feasible space). Second, it is unclear how to realistically sample the contents of short-term memory. Consequently, we are currently forced to use $TS_{N1}$ to generate, via a Monte Carlo-like process, a representative set of samples. Further, we note that the often deterministic behavior of $TS_{N1}$ (discounting ties in the case of multiple equally good non-tabu moves and randomization of the tabu tenure) generally prevents direct characterization of the *distribution* of transition probabilities for any *single* sample, as is possible for local search algorithms with a stronger stochastic component, e.g., iterated local search or Metropolis sampling (Watson, 2003).

In Figure 5, we show the estimated probabilities of moving closer to (left figure) or farther from (right figure) the nearest optimal solution for a typical $10 \times 10$ random JSP; the probability of maintaining an *equal* search gradient is negligible ($p < 0.1$), independent of the current distance to the nearest optimal solution. We observe qualitatively similar results for all of our $6 \times 4$, $6 \times 6$, and $10 \times 10$ random JSPs, although we note that results for most instances generally possess more noise (i.e., small-scale irregularities) than those observed in Figure 5. The results indicate that the probability of continuing to move closer to (farther from) the nearest optimal solution is typically proportional (inversely proportional) to the current distance from the nearest optimum. An exception occurs when $i \leq 10$ and the gradient is *closer*, where the probability of continuing to move closer to an optimal solution actually rises as $i \to 0$. We currently have no explanation for this phenomenon, although it appears to be due in part to the steepest-descent bias exhibited by $TS_{N1}$.

The probabilities of moving closer to/farther from the nearest optimal solution are, in general, roughly symmetric around $D_{max}/2$, such that search in $TS_{N1}$ is biased toward solutions that are an average distance from the nearest optimal solution. This characteristic provides an explanation for the Gaussian-like distributions of $d_{opt}$ observed for solutions visited during search, e.g., as shown in Figure 2. The impact of short-term memory is also evident, as the probability of maintaining the current search gradient is high and consistently exceeds 0.5 in all of the problem instances we examined, with the exception of occasional brief drops to no lower than 0.4 at extremal distances $i$, i.e., $i \approx 0$ or $i \approx D_{max}$. The probability of inverting the current gradient is also a function of the distance to the nearest optimal solution and the degree of change. For example, the probability of switching gradients from *equal* to *closer* is higher than the probability of switching from *farther* to *closer*. Consistent with the results presented above in Section 6, (1) the distance $i$ at which $P(S_{i-1,closer}|S_{i,closer}) = P(S_{i+1,farther}|S_{i,farther})$ is approximately equal to $\overline{d}_{tabu\text{-}opt}$ and (2) $\overline{d}_{lopt\text{-}opt}$ generally falls anywhere in the range $[D_{max}/2, D_{max}]$. Finally, we note the resemblance between the transition probabilities in our Markov model and those in the well-known Ehrenfest model found in the literature on probability theory (Feller, 1968, p. 377); in both models, the random walk dynamics can be viewed as a simple diffusion process with a central restoring force.

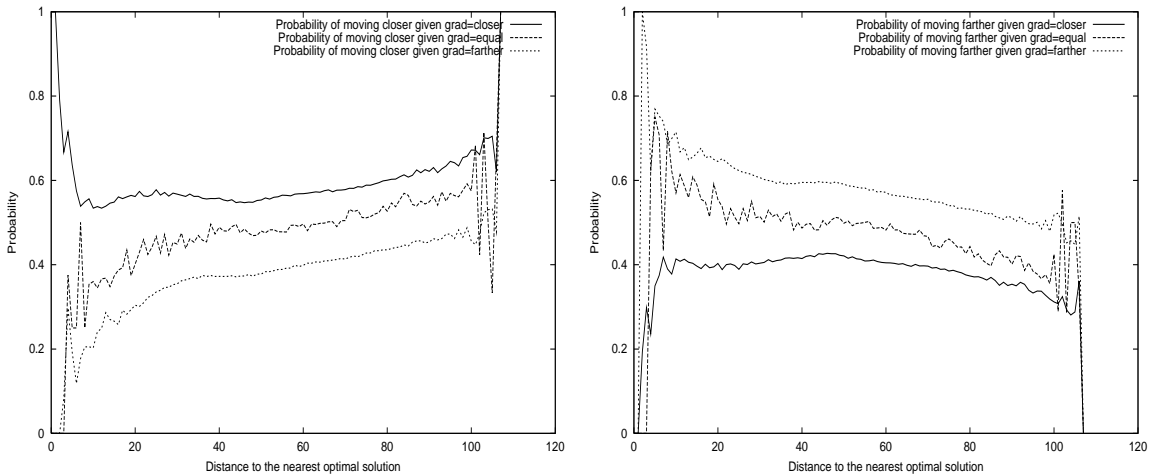

Figure 5: The transition probabilities for moving closer to (left figure) or farther from (right figure) the nearest optimal solution under $TS_{N1}$ for a typical $10 \times 10$ random JSP.

### 7.3 Validation

To validate the random walk model, we compare the actual mean search cost $\overline{c}$ observed under $TS_{N1}$ with the corresponding value predicted by the model. We then construct a $log_{10}$-$log_{10}$ linear regression model of the predicted versus actual $\overline{c}$ and quantify model accuracy as the resulting $r^2$. Because it is based on the random walk model of $TS_{N1}$, we refer to the resulting linear regression model as a *dynamic* cost model. Due to the close relationship between the random walk and dynamic cost models, we use the two terms interchangeably when identification of a more specific context is unnecessary.

To compute the predicted $\overline{c}$ for a given problem instance, we repeatedly simulate the corresponding random walk model defined by the parameters $D_{max}$, the set of states $S_{i,x}$, and the estimated transition probabilities $P(S_{j,x'}|S_{i,x})$. Each simulation trial is initiated from a state $S_{m,n}$, where $m = d_{opt}(s)$ for a trial-specific random local optimum $s$ and $n$ equals *closer* or *farther* with equal probability; recall that the probability of maintaining an *equal* search gradient is negligible. We compute $m$ exactly (as opposed to simply using $rint(\overline{d}_{lopt\text{-}opt})$) in order to control for possible effects of the distribution of $d_{opt}$ for random local optima, which tend to be more irregular (i.e., non-Gaussian) for small problem instances; letting $m = rint(\overline{d}_{lopt\text{-}opt})$ results in a slight ($< 5\%$) decrease in model accuracy. We then define the predicted search cost $\overline{c}$ as the mean number of simulated iterations required to achieve the absorbing state $S_{0,closer}$; statistics are taken over 10,000 independent trials.

We first consider results obtained for our $6 \times 4$ and $6 \times 6$ random JSPs. Scatter-plots of the predicted versus actual $\overline{c}$ for these two problem sets are shown in the top portion of Figure 6. The $r^2$ value for both of the corresponding $log_{10}$-$log_{10}$ regression models is a remarkable 0.96. For all but 21 and 11 of the respective 1,000 $6 \times 4$ and $6 \times 6$ instances, the actual $\overline{c}$ is within a factor of 2 of the predicted $\overline{c}$. For the remaining instances, the actual $\overline{c}$ deviates from the predicted value by a maximum factor of 4.5 and 3.5, respectively. In contrast to the $\overline{d}_{lopt\text{-}opt}$ and $\overline{d}_{tabu\text{-}opt}$ cost models, there is no evidence of an inverse correlation between problem difficulty and model accuracy; if anything, the model is least accurate for the *easiest* problem instances, as shown in the upper left side of Figure 6. A detailed examination of the high-residual instances indicates that the source of the prediction error is generally the fact that $TS_{N1}$ visits specific subsets of solutions that are close to optimal solutions with a disproportionately high frequency, such that the primary assumption underlying our Markov model, i.e, lumpability, is grossly violated. As shown below, we have not yet observed this behavior in sets of larger random JSPs, raising the possibility that the phenomena is isolated.

Next, we assess the scalability of the dynamic cost model by considering the 42 of our $10 \times 10$ random JSPs with $\leq 100,000$ optimal solutions. A scatter-plot of the predicted versus actual $\overline{c}$ for these instances is shown in the lower portion of Figure 6; the $r^2$ value of the corresponding $log_{10}$-$log_{10}$ regression model is 0.97. For reference, we include results (labeled) for those $10 \times 10$ random JSPs from the OR Library with $\leq 100,000$ optimal solutions. The actual $\overline{c}$ is always within a factor of 2.1 of the predicted $\overline{c}$, and there is no evidence of any correlation between accuracy and problem difficulty. More importantly, we observe no degradation in accuracy relative to the smaller problem sets.

We have also explored a number of secondary criteria for validation of the dynamic cost model. In particular, we observe minimal differences between the predicted and actual statistical *distributions* of both (1) the distances to the nearest optimal solution and (2) the trend lengths, i.e., the number of iterations that consistent search gradients are maintained. Additionally, we consider differences in the distribution of predicted versus actual search costs below in Section 9. Finally, we note that the dynamic cost model is equally accurate ($r^2 \geq 0.96$) in accounting for the cost of locating sub-optimal solutions to arbitrary $6 \times 4$ and $6 \times 6$ random JSPs, as well as specially constructed sets of very difficult $6 \times 4$ and $6 \times 6$ random JSPs. Both problem types are fully detailed by Watson (2003) .

### 7.4 Discussion

The results presented in this section provide strong, direct evidence for our hypothesis that search under $TS_{N1}$ acts as a variant of a straightforward one-dimensional random walk over the $S_{lopt+}$ sub-space. However, the transition probabilities between states of the random walk are non-uniform, reflecting the presence of two specific biases in the search dynamics. First, search is biased toward solutions that are approximately equi-distant from the nearest optimal solution and solutions that are maximally distant from the nearest optimal solution. Consequently, in terms of random walk theory, the run-time dynamics can be viewed as a diffusion process with a central restoring force toward solutions that are an average distance from the nearest optimal solution. Second, $TS_{N1}$'s short-term memory causes search to consistently progress either toward or away from the nearest optimal solution for extended time periods; such strong trending behavior has not been observed in random walks or in other memoryless local search algorithms for the JSP (Watson, 2003). Despite its central role in tabu search, our analysis indicates that, surprisingly, short-term memory is not always beneficial. If search is progressing toward an optimal solution, then short-term memory will increase the probability that search will proceed even closer. In contrast, when search is moving away from an optimal solution, short-term memory inflates the probability that search will continue to be led astray. Finally, we note that like $\bar{d}_{lopt\text{-}opt}$ and $\bar{d}_{tabu\text{-}opt}$, $D_{max}$ is a concrete measure of $|S_{lopt+}|'$; all three measures directly quantify, with varying degrees of accuracy, the *width* of the search space explored by $TS_{N1}$. We further discuss the linkage between these measures below in Section 8.

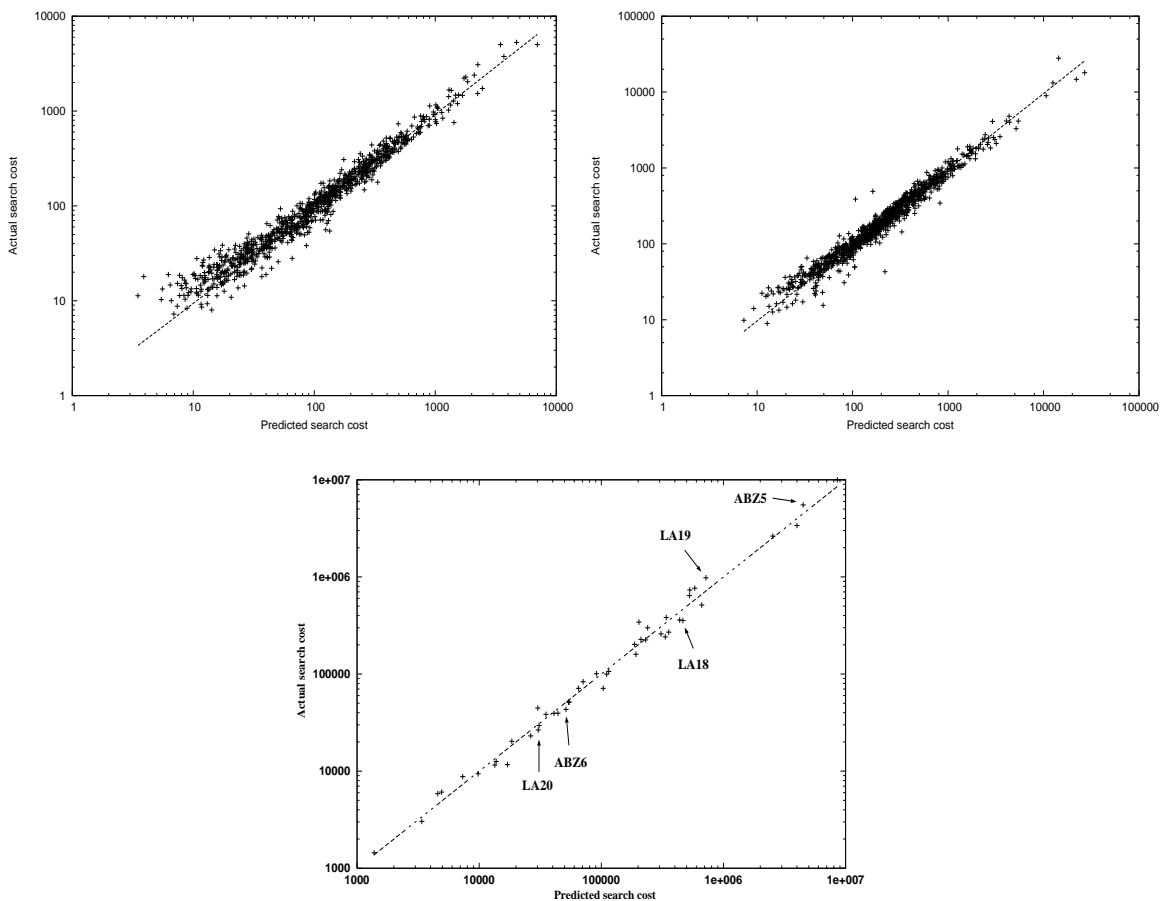

Figure 6: Scatter-plots of the predicted versus actual search cost $\bar{c}$ for $6 \times 4$ (upper left figure), $6 \times 6$ (upper right figure), and $10 \times 10$ (lower figure) random JSPs; the least-squares fit lines are super-imposed.

### 7.5 Related Research

Hoos (2002) uses Markov models similar to those presented here to analyze the source of specific irregularities observed in the run-length distributions (see Section 9) of some local search algorithms for SAT. Because the particular algorithms investigated by Hoos are memoryless, states in the corresponding Markov chain model simply represent the set of solutions distance $k$ from the nearest optimal (or more appropriately in the case of SAT, satisfying) solution. The transition probabilities for moving either closer to or farther from an optimal solution are fixed to the respective constant values $p^-$ and $p^+ = 1 - p^-$, independent of $k$. By varying the values of $p^-$, Hoos demonstrates that the resulting Markov chains exhibit the same types of run-length distributions as well-known local search algorithms for SAT, including GWSAT and WalkSAT. Extensions of this model are additionally used to analyze stagnation behavior that is occasionally exhibited by these same algorithms.

Our research differs from that of Hoos in several respects, the most obvious of which is the explicit modeling of $TS_{N1}$'s short-term memory mechanism. More importantly, we derive estimates of both the transition probabilities and the number of states directly from instance-specific data. We then test the ability of the resulting model to capture the behavior of $TS_{N1}$ on the specific problem instance. In contrast, Hoos posits a particular structure to the transition probabilities *a priori*. Then, by varying parameter values such as $p^-$ and the number of model states, Hoos demonstrates that the resulting models capture the range of run-length distributions exhibited by local search algorithms for SAT; accuracy relative to individual instances is not assessed.

Additionally, we found no evidence that the transition probabilities in the JSP are independent of the current distance to the nearest optimal solution. Given that (1) the solution representation underlying $TS_{N1}$ and many other local search algorithms for the JSP is a Binary hypercube and (2) neighbors under the *N1* operator are by definition Hamming distance 1 from the current solution, constant transition probabilities would be entirely unexpected from a theoretical standpoint (Watson, 2003). Finally, we have developed analogous dynamic cost models for a number of memoryless local search algorithms for the JSP based on the *N1* move operator, including a pure random walk, iterated local search, and Metropolis sampling (Watson, 2003).

Finally, there are similarities between our notion of effective search space size ($|S_{lopt+}|'$) and the concept of a *virtual* search space size. Hoos (1998) observes that local search algorithms exhibiting exponentially distributed search costs (which includes $TS_{N1}$, as discussed in Section 9) behave in a manner identical to blind guessing in a sub-space of solutions containing both globally optimal and sub-optimal solutions. Under this interpretation, more effective local search algorithms are able to restrict the total number of sub-optimal solutions under consideration, i.e., they operate in a smaller virtual search space. Our notion of effective search space size captures a similar intuition, but is in contrast grounded directly in terms of search space analysis; in effect, we provide an answer to a question posed by Hoos, who indicates "ideally, we would like to be able to identify structural features of the original search space which can be shown to be tightly correlated with virtual search space size" (1998, p. 133). Further, we emphasize the role of the number *and* distribution of globally optimal optimal solutions within the sub-space of solutions under consideration, and directly relate run-time dynamics (as opposed to search cost distributions) to effective search space size (i.e., through $D_{max}$).

## 8. The Link Between Search Space Structure and Run-Time Dynamics

In transitioning from static to dynamic cost models, our focus shifted from algorithm-independent features of the fitness landscape to explicit models of algorithm run-time behavior. By leveraging increasingly detailed information, we were able to obtain monotonic improvements in cost model accuracy. Discounting the difficulties associated with identifying the appropriate types of information, a positive correlation between information quantity and cost model accuracy is conceptually unsurprising. Static cost models are algorithm-independent – a useful feature in certain contexts – and we anticipate a weak upper bound on the absolute accuracy of these models. Our quasi-dynamic $\overline{d}_{tabu\text{-}opt}$ cost model is based on the same summary statistic as the static $\overline{d}_{lopt\text{-}opt}$ cost model; only the sample sets involved in computation of the statistic are different. In either case, the resulting cost models are surprisingly accurate, especially given the simplicity of the statistic. In contrast, a comparatively overwhelming increase in the amount of information appears to be required (as embodied in the dynamic cost model) to achieve further increases in accuracy.

Perhaps more interesting than the correlation between information quantity and cost model accuracy is the nature of the relationship between the information underlying cost models at successive "levels", i.e., between static and quasi-dynamic models, or quasi-dynamic and dynamic models. Specifically, we argue that the parameters associated with a particular cost model estimate key parameters of the cost model at the subsequent higher level, exposing an unexpectedly strong and simple link between fitness landscape structure in the JSP and the run-time dynamics of $TS_{N1}$.

Recall from Section 7.2 that the estimated transition probabilities in the dynamic cost model are qualitatively identical across the range of problem instances and that most major differences are due to variability in $D_{max}$, the maximal observed distance to the nearest optimal solution. Further, we observe that the "closer" and "farther" transition probabilities are roughly symmetric around $D_{max}/2$. Consequently, search under $TS_{N1}$ is necessarily biased toward solutions that are approximately distance $D_{max}/2$ from the nearest optimal solution. More precisely, we denote the mean predicted distance to the nearest optimal solution by $\overline{d}_{dynamic\text{-}opt}$; simulation confirms that $\overline{d}_{dynamic\text{-}opt} \approx D_{max}/2$, where any deviations are due primarily to the asymmetric rise in transition probability as $i \to 0$ for gradients equal to *closer*. But $\overline{d}_{tabu\text{-}opt}$ also measures the mean distance between solutions visited during search and the nearest optimal solution, i.e., $\overline{d}_{tabu\text{-}opt} \approx \overline{d}_{dynamic\text{-}opt}$. Thus, we believe the success of the $\overline{d}_{tabu\text{-}opt}$ model is due to the fact that (1) the transition probabilities in the dynamic cost model are qualitatively identical across different problem instances and (2) $\overline{d}_{tabu\text{-}opt}$ indirectly approximates the key parameter $D_{max}$ of the dynamic cost model, via the relation $\overline{d}_{tabu\text{-}opt} \approx \overline{d}_{dynamic\text{-}opt} \approx D_{max}/2$. Discrepancies in the accuracy of the dynamic and quasi-dynamic cost models are expected, as no single measure is likely to capture the impact of subtle irregularities in the transition probabilities. The power of the $\overline{d}_{lopt\text{-}opt}$ model is in turn due to the fact that $\overline{d}_{lopt\text{-}opt} \approx \overline{d}_{tabu\text{-}opt}$ – but only for small problem instances. For larger problem instances, $\overline{d}_{lopt\text{-}opt}$ consistently over-estimates $\overline{d}_{tabu\text{-}opt}$, and consequently $D_{max}$, by failing to discount those regions of the search space that $TS_{N1}$ is unlikely to explore. To conclude, the various cost models are related by the expression $\overline{d}_{lopt\text{-}opt} \approx \overline{d}_{tabu\text{-}opt} \approx \overline{d}_{dynamic\text{-}opt} \approx D_{max}/2$. The linkage between the models is due to the fact that these models all attempt, with varying degrees of accuracy, to quantify the effective size $|S_{lopt+}|'$ of the sub-space of solutions likely to be visited by $TS_{N1}$ during search.

## 9. The Dynamic Cost Model and Run-Length Distributions

The cost models developed in Sections 5–7 account for variability in central tendency measures of problem difficulty, i.e., $\overline{c}$ and $c_{Q2}$. In reality, search cost is a random variable $C$. Consequently, a cost model should ideally both qualitatively and quantitatively capture the full *distribution* of $C$. Because they are based on simple summary statistics, it is difficult to imagine how static and quasi-dynamic cost models might be extended to account for $C$. In contrast, a predicted $C$ is easily obtained from the dynamic cost model; as discussed in Section 7.3, a predicted $C$ is generated in order to compute $\overline{c}$ and is subsequently discarded. We now analyze the nature of the full $C$ predicted by the dynamic cost model and determine whether it accurately represents the actual distribution of search costs under $TS_{N1}$.

We follow Hoos (1998) in referring to the distribution $C$ for a given problem instance as the *run-length distribution* (RLD). In what follows, we consider two specific questions relating to the RLDs of random JSPs under $TS_{N1}$: (1) From what family of distributions are the RLDs drawn? and (2) Are the predicted and actual RLDs identically distributed? Both questions can be answered using standard statistical goodness-of-fit tests. Although the RLDs for $TS_{N1}$ are discrete, we approximate the actual distributions using continuous random variables; the approximation is tight due to the wide range of search costs observed across individual trials, allowing us to avoid issues related to the specification of the bin size in the standard $\chi^2$ goodness-of-fit test for discrete random variables (e.g., such as those performed by Hoos). Instead, we use the two-sample Kolmogorov-Smirnov (KS) goodness-of-fit test, which assumes the existence of samples (in the form of cumulative density functions or CDFs) for distinct continuous random variables. The null hypothesis for the KS test states that the distributions underlying both samples are identically distributed. The KS test statistic quantifies the maximal distance between the two CDFs and the null hypothesis is rejected if the "distance" between them is sufficiently large (Scheaffer & McClave, 1990).

We first consider the family of distributions from which the individual RLDs are drawn. Taillard (1994, p. 116) indicates that the number of iterations required to locate optimal solutions using his algorithm is approximately exponentially distributed. However, he only reported qualitative results for a single $10 \times 10$ problem instance. Using our set of $10 \times 10$ random JSPs, we now perform a more comprehensive analysis. For each instance, we compute the two-sample KS test statistic for the null hypothesis that the RLD is exponentially distributed. The first sample consists of the actual search costs $c$ observed over 1,000 independent trials. The second sample consists of 1,000,000 random samples drawn from an exponential distribution with a mean $\overline{c}$ computed from the first sample; direct sampling from the theoretical CDF is required by the particular statistical software package we employed in our analysis. We show the distribution of the p-values associated with the resulting KS test statistics in the left side of Figure 7. At $p \leq 0.01$, we reject the null hypothesis for 11 of the 100 instances, i.e., search cost under $TS_{N1}$ is *not* exponentially distributed in roughly 10% of the instances. In the right side of Figure 7, we show CDFs of both the actual RLD and the corresponding exponential RLD for the instance with the smallest p-value, or, equivalently, the largest deviation between the two CDFs. For this instance, and all other instances with $p \leq 0.01$, the two distributions differ primarily in their left tails. In particular, we observe far fewer low-cost runs than found in a purely exponential distribution.

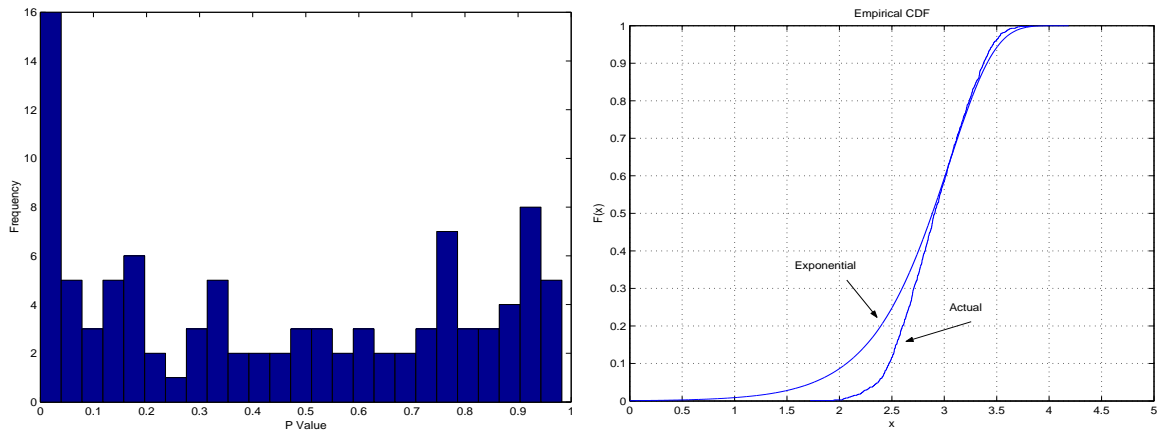

Figure 7: Left Figure: Distribution of the p-values for rejecting the null hypothesis that the RLDs of $10 \times 10$ random JSPs are exponentially distributed under $TS_{N1}$. Right Figure: The actual and exponential RLDs for the $10 \times 10$ instance with the smallest p-value.

Our results reinforce Taillard's observation that the RLDs under $TS_{N1}$ are approximately exponentially distributed. Exponential RLDs also arise in the context of local search algorithms for other $NP$-hard problems. For example, Hoos (1998, p. 118) reports qualitatively similar results for a range of local search algorithms (e.g., Walk-SAT) for MAX-SAT. Hoos additionally demonstrated that the deviation from the exponential "ideal" is a function of problem difficulty: RLDs of harder (easier) problem instances are more (less) accurately modeled by an exponential distribution. A similar relationship holds for the RLDs under $TS_{N1}$. In Figure 8, we show a scatter-plot of the mean search cost $\bar{c}$ versus the value of the KS test statistic for $10 \times 10$ random JSPs. The data indicate that the value of the KS test statistic is inversely proportional to instance difficulty. More specifically, the RLDs under $TS_{N1}$ are approximately exponential for moderate-to-difficult instances, while the exponential approximation degrades for easier instances, e.g., as shown in the right side of Figure 7. Given significant differences between MAX-SAT and the JSP, our result raises the possibility of a more universal phenomenon. Finally, we note that Hoos also demonstrated that the RLDs of easy instances are well-approximated by a Weibull distribution, a generalization of the exponential distribution. Although not reported here, this finding also translates to the RLDs of those instances shown in Figure 8 with $p \leq 0.05$.

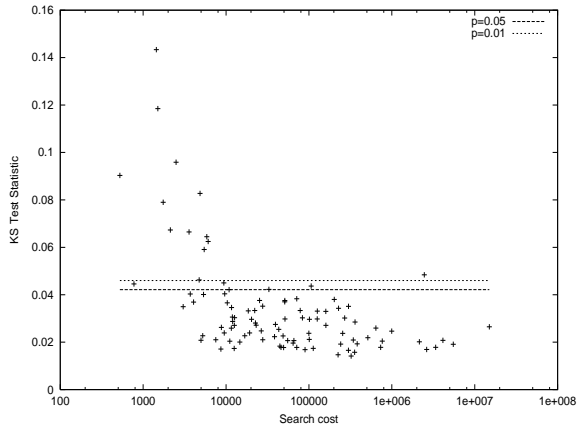

Figure 8: Scatter-plot of mean search cost ($\bar{c}$) versus the value of the Kolmogorov-Smirnov test statistic for comparing the actual search cost distribution with that of the corresponding exponential distribution. Large values of the test statistic indicate more significant differences. The horizontal lines indicate null hypothesis rejection thresholds at $p \leq 0.01$ and $p \leq 0.05$.

Next, we analyze whether the RLDs predicted by the dynamic cost model can account for the actual RLDs observed for $TS_{N1}$. For each of our $10 \times 10$ random JSPs, we compute the two-sample KS test statistic for the null hypothesis that the predicted and actual RLDs originate from the same underlying distribution. The first sample consists of the actual $c$ observed over 1,000 independent trials of $TS_{N1}$. The second sample consists of the 10,000 individual costs $c$ used to estimate the predicted $\bar{c}$ (see Section 7.3). The discrepancy in the two sample sizes stems from the cost associated with obtaining individual samples. We only report results for the 42 instances for which estimation of the dynamic cost model parameters is computationally tractable. For all but 6 of the 42 instances ($\leq 15\%$), we reject the null hypothesis that the two distributions are identical at $p \leq 0.01$; we found no evidence of any correlation between $p$ and problem difficulty. In other words, despite the success of the dynamic cost model in accounting for $\bar{c}$, it generally fails to account for the full RLD. Yet, despite statistically significant differences, the predicted and actual RLDs are generally qualitatively identical. For example, consider the predicted and actual RLDs for the two instances yielding the smallest and largest p-values, as shown in Figure 9. In the best case, the two distributions are effectively identical. In the *worst* case, the two distributions appear to be qualitatively identical, such that the actual RLD can be closely approximated by shifting the predicted RLD along the x-axis.

In further support of this observation, we more carefully consider the 36 instances in which the differences between the actual and predicted RLDs are statistically significant at $p \leq 0.01$. For each instance, we compute the KS test statistic for the difference between the predicted RLD and an exponential distribution with mean equal to the predicted $\bar{c}$. For all difficult instances, specifically those with a predicted $\bar{c} \geq 10,000$, we fail to reject the null hypothesis that the underlying distributions are identical at $p \leq 0.01$. Consequently, if the dynamic cost model were able to accurately predict $\bar{c}$, the predicted and actual RLDs would be statistically indistinguishable. For the remaining easy and medium instances, we do observe statistically significant differences at $p \leq 0.01$ between the predicted and corresponding exponential RLDs. However, any differences are isolated to the left tails of both distributions, such that the predicted RLDs under-cut the corresponding exponential RLDs, e.g., as shown in the right side of Figure 7. In other words, the RLDs predicted by the dynamic cost model capture those deviations from the exponential form that are observed in the RLDs of $TS_{N1}$ on easy and medium difficulty instances.

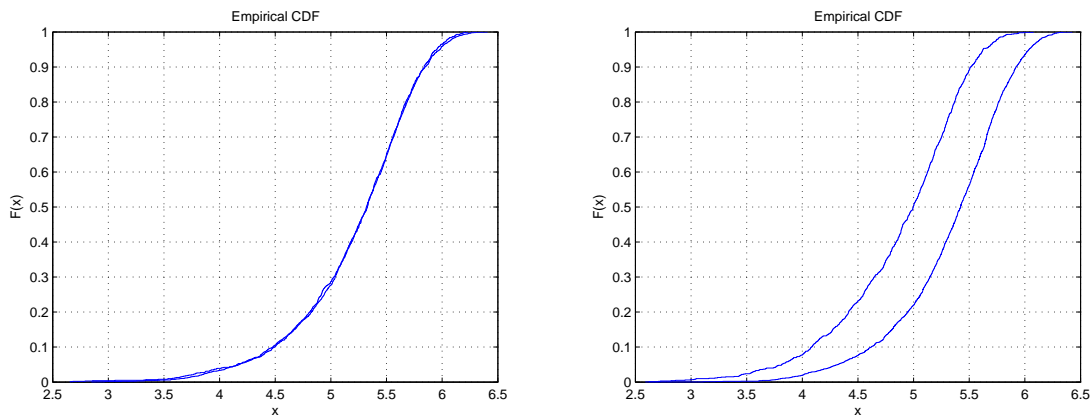

Figure 9: CDFs of the predicted and actual RLDs for two $10 \times 10$ random JSPs. The p-values for the KS test statistic are respectively 0.73 and $3.4 \times 10^{-78}$.

Given (1) the well-known difficulty of accurately estimating the mean of an exponential or exponential-like distribution and (2) the fact that any "lumped" model of $TS_{N1}$ will necessarily fail to capture the full detail of the true underlying Markov chain, our results provide strong evidence in support of the hypothesis that any observed differences between the predicted and actual RLDs are not indicative of any major structural flaw in the dynamic cost model.

## 10. On the Difficulty of Structured JSPs

For fixed $n$ and $m$, the mean difficulty of JSP instances is known to increase as the number of workflow partitions *wf* is varied from 1 (corresponding to random JSPs) to $m$ (corresponding to flowshop JSPs), i.e., as more structure is introduced. For example, the mean $c_{Q2}$ under $TS_{N1}$ observed for our $6 \times 6$ random, workflow, and flowshop JSPs are 280, 3,137, and 12,127, respectively. These differences represent an order-of-magnitude increase in average difficulty as $wf$ is varied from 1 to $m/2$ and again from $m/2$ to $m$. Similar differences are observed for $10 \times 10$ random, workflow, and flowshop JSPs, where the mean $c_{Q2}$ are respectively 315,413, $4.35 \times 10^7$, and $2.62 \times 10^8$. The often extreme difficulty of structured JSPs is further illustrated by the fact that the most difficult $10 \times 10$ flowshop JSP required an average of 900 *million* iterations of $TS_{N1}$ to locate an optimal solution.

|  |  | Cost Model | |
|---|---|---|---|
| Problem Size | Structure | $\overline{d}_{lopt\text{-}opt}$ | $\overline{d}_{tabu\text{-}opt}$ |
| $6 \times 4$ | Random | 0.80 | 0.84 |
|  | Workflow | 0.62 | 0.76 |
|  | Flowshop | 0.70 | 0.72 |
| $6 \times 6$ | Random | 0.64 | 0.78 |
|  | Workflow | 0.30 | 0.55 |
|  | Flowshop | 0.41 | 0.55 |

Table 2: The $r^2$ values of the $\overline{d}_{lopt\text{-}opt}$ and $\overline{d}_{tabu\text{-}opt}$ cost models of $TS_{N1}$ obtained for $6 \times 4$ and $6 \times 6$ random, workflow, and flowshop JSPs.

We previously reported that the accuracy of the $\overline{d}_{lopt\text{-}opt}$ cost model fails to transfer from random JSPs to workflow JSPs (Watson et al., 2003); additional experiments yield similar results for flowshop JSPs. Table 2 shows the $r^2$ values of the $\overline{d}_{lopt\text{-}opt}$ and $\overline{d}_{tabu\text{-}opt}$ cost models for $6 \times 4$ and $6 \times 6$ random, workflow, and flowshop JSPs. The results indicate that the $\overline{d}_{lopt\text{-}opt}$ model is more accurate on random JSPs than on either workflow and flowshop JSPs, although accuracy improves when transitioning from workflow to flowshop JSPs. The $\overline{d}_{tabu\text{-}opt}$ model is more accurate than the $\overline{d}_{lopt\text{-}opt}$ model on both types of structured JSP. However, despite the absolute improvements relative to the $\overline{d}_{lopt\text{-}opt}$ model, accuracy of the $\overline{d}_{tabu\text{-}opt}$ model decreases with increases in $wf$. Overall, the $\overline{d}_{tabu\text{-}opt}$ model accounts for slightly over half of the variability in problem difficulty observed in the more difficult structured $6 \times 6$ JSPs. The significant difference in accuracy ($\approx 30\%$) relative to random JSPs raises the possibility that the dynamic cost model may be unable to correct for the deficiencies of the $\overline{d}_{tabu\text{-}opt}$ model. In particular, factors other than $|S_{lopt+}|'$ may be contributing to the difficulty of structured JSPs, or it may not be possible to model the behavior of $TS_{N1}$ on structured instances as a simple random walk.

Consider the transition probabilities under the dynamic cost model for $6 \times 4$ and $6 \times 6$ workflow and flowshop JSPs, obtained using the sampling methodology described in Section 7.2. Figure 10 shows the probabilities of $TS_{N1}$ continuing to transition closer to the nearest optimal solution for two $6 \times 6$ flowshop instances. For the instance corresponding to the left-hand figure, the transition probabilities are roughly proportional to the current distance from the nearest optimal solution, which is consistent with the results observed for random JSPs, e.g., as shown earlier in Figure 5. In contrast, for the instance corresponding to the right-hand figure, the probability of $TS_{N1}$ continuing to move closer to the nearest optimal solution is effectively constant at $\approx 0.6$. These examples illustrate a key point regarding the behavior of $TS_{N1}$ on structured JSPs: the transition probabilities are significantly more heterogeneous than those observed for random JSPs, often deviating significantly from the "prototypical" (i.e., symmetric around $D_{max}/2$) form in *both* qualitative and quantitative aspects. Given such large deviations, it is unsurprising that cost models based on simple summary statistics, specifically $\overline{d}_{tabu\text{-}opt}$, fail to account for a substantial proportion of the variability in the difficulty of structured JSPs.

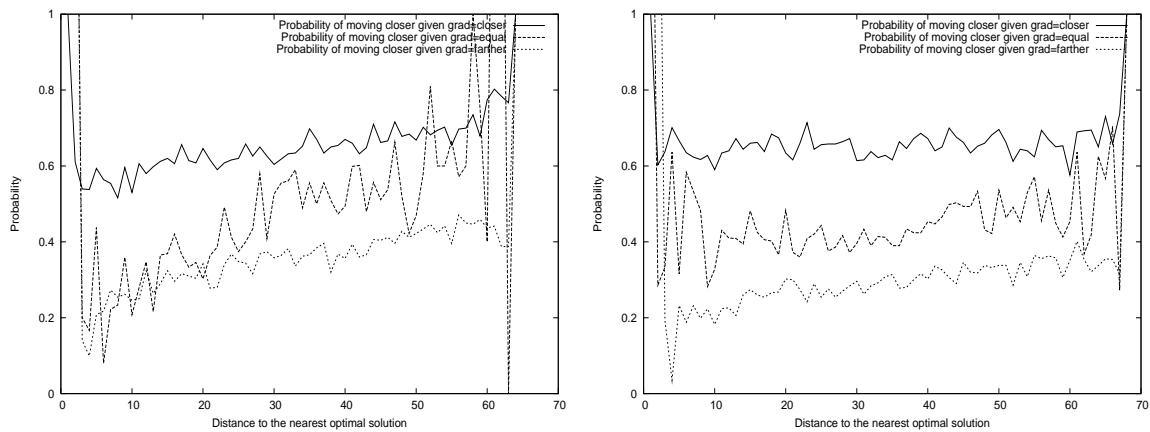

Figure 10: The transition probabilities for moving closer to the nearest optimal solution under $TS_{N1}$ for distinct $6 \times 6$ flowshop JSPs.

Despite differences in the transition probabilities relative to those observed for random JSPs, we observe no impact on the overall accuracy of the dynamic cost model. For $6 \times 4$ and $6 \times 6$ workflow JSPs, the $r^2$ values corresponding to a $log_{10} - log_{10}$ regression of the predicted versus actual $\overline{c}$ are respectively 0.97 and 0.95. The analogous $r^2$ for both $6 \times 4$ and $6 \times 6$ flowshop JSPs is 0.96. For all but a few exceptional instances, the actual $\overline{c}$ is always within a factor of 3 of the predicted $\overline{c}$; in no case does the difference exceed a factor of 4. Similar results are obtained on a small set of randomly generated $10 \times 10$ workflow and flowshop JSPs for which transition probability estimation is computationally feasible.

The dynamic cost model accurately captures the run-time dynamics of $TS_{N1}$ on *both* random and structured JSPs, although the transition probabilities are qualitatively different in the two types of problem. Heterogeneity in the transition probabilities of structured JSPs additionally indicates that variability in the difficulty of these instances is unlikely to be captured by simple summary statistics, yielding reductions in the accuracy of our static and quasi-dynamic cost models. Finally, we observe that it is still possible that the accuracy of the dynamic cost model may degrade significantly under fundamentally different types of structure than those considered here, e.g., with strong correlation between subsets of the operation durations $\tau_{ij}$.

Mattfeld et al. (1999) indicate that differences in the difficulty of random and workflow JSPs are due to differences in the size of the search spaces, as measured by the mean distance between random local optima. The observed mean maximal distance $D_{max}$ to a nearest optimal solution increases when moving from random JSPs to workflow JSPs, and again from workflow JSPs to flowshop JSPs; results for $6 \times 4$ and $6 \times 6$ problem sets are shown in Table 3. Consequently, our results serve to clarify Mattfeld et al.'s original assertion: given that $D_{max}$ is a measure of $|S_{lopt+}|'$, differences in the difficulty of random and structured JSPs are simply due to differences in the effective size of the search space.

|  | Structure Type | | |
|---|---|---|---|
| Problem Size | Random | Workflow | Flowshop |
| $6 \times 4$ | 21.46 | 37.01 | 44.80 |
| $6 \times 6$ | 26.67 | 47.80 | 68.86 |

Table 3: The mean maximal distance to the nearest optimal solution ($D_{max}$) observed for $6 \times 4$ and $6 \times 6$ random, workflow, and flowshop JSPs.

## 11. Moving Toward Cost Models of State-of-the-Art Algorithms

As discussed in Section 3, $TS_{N1}$ is a relatively straightforward implementation of tabu search for the JSP. In particular, it lacks features such as advanced move operators and long-term memory mechanisms that have been demonstrated to improve the performance of tabu search algorithms for the JSP. Given an accurate cost model of $TS_{N1}$, the next logical step is to systematically assess the impact of these algorithmic features on model structure and accuracy. Ultimately, the goal is to incrementally move both the target algorithm and the associated cost model toward the state-of-the-art, e.g., as currently represented by Nowicki and Smutnicki's (2005) *i*-TSAB algorithm. We now take an initial step toward this goal by demonstrating that a key performance-enhancing component – the powerful *N5* move operator – fails to impact either the structure or accuracy of the dynamic cost model.

Recall from Section 3 that the neighborhood of a solution $s$ under the *N1* move operator consists of all solutions obtained by inverting the order of a pair of adjacent operations on the same critical block. Let $s' \in N1(s)$ denote the solution obtained by inverting the order of two adjacent critical operations $o_{ij}$ and $o_{kl}$ in $s$. If both $o_{ij}$ and $o_{kl}$ are contained entirely within a critical block, i.e., neither operation appears first or last in the block, then $C_{max}(s') \geq C_{max}(s)$ (Mattfeld, 1996). In other words, many moves under *N1* provably cannot yield immediate improvements in the makespan of the current solution and therefore should not be considered during search. Building on this observation, Nowicki and Smutnicki (1996) introduce a highly restricted variant of the *N1* move operator, with the goal of accelerating local search by reducing the total cost of neighborhood evaluation. This operator, which we denote *N5*, contributes significantly to the performance of both the well-known TSAB algorithm (Nowicki & Smutnicki, 1996) and the current state-of-the-art algorithm for the JSP, *i*-TSAB (Nowicki & Smutnicki, 2005). However, the power of the *N5* operator comes with a price: the induced search space is *disconnected*, such that it is not always possible to move from an arbitrary solution to a globally optimal solution. Consequently, no local search algorithm based strictly on *N5* can be PAC in the theoretical sense. Further, as discussed below, empirical PAC behavior for basic tabu search algorithms based on the *N5* operator is not possible for random JSPs.

The lack of the PAC property significantly complicates the development of cost models, as it is unclear how to quantify search cost or, equivalently, problem difficulty. This fact, in large part, drove our decision to base our research on the $TS_{N1}$ algorithm. Recently, we demonstrated that for random JSPs the *N5* operator can induce small "traps" or isolated sub-components of the search space from which escape is impossible (Watson, 2003). Fortunately, these traps are easily detected and can be escaped via a short random walk under the more general (i.e., connected) $N1$ move operator. Given these observations, we introduced a novel $N5$-based tabu search algorithm for the JSP that is empirically PAC on all of our $6 \times 4$, $6 \times 6$, and $10 \times 10$ sets of random JSPs. The algorithm, which we denote $TS_{N5}$, is detailed by Watson (2003). However, with the exception of both the move operator and the trap detection/escape mechanisms, $TS_{N1}$ and $TS_{N5}$ are identical. We further observe that traps are infrequently encountered (typically at most once every 5K or more iterations), and we use identical settings for all parameters found in both algorithms, i.e., $L_{min}$ and $L_{max}$. Consequently, we are now able to characterize the impact of the $N5$ move operator on the effectiveness of tabu search for the JSP under controlled experimental conditions.

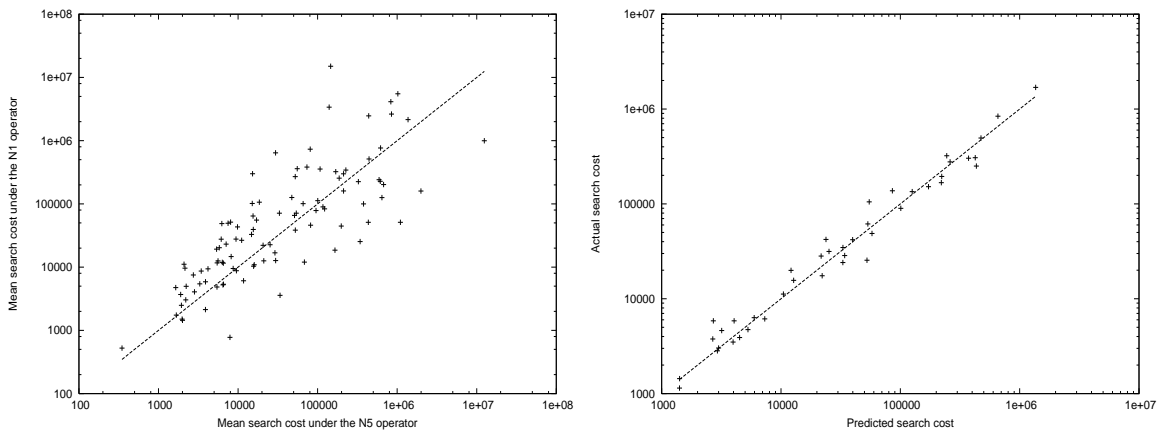

Figure 11: Left Figure: Scatter-plot of the actual search cost $\bar{c}$ under the $TS_{N5}$ versus $TS_{N1}$ algorithms; the line $y = x$ is super-imposed. Right Figure: Scatter-plot of the predicted versus actual search cost $\bar{c}$ for $10 \times 10$ random JSPs using $TS_{N5}$; the least-squares fit line is super-imposed.

Consider the mean search cost $\bar{c}$ under both $TS_{N1}$ and $TS_{N5}$ on our $10 \times 10$ random JSPs; the corresponding scatter-plot is shown in the left side of Figure 11. We observe unexpectedly low correlation between problem difficulty under the two algorithms; differences of a factor of 10 are common and reach nearly a factor of 100 in the worst case. Due to the low frequency of occurrence, a minimal proportion of the observed differences can be attributed to the trap detection and escape mechanisms. The implication is that the move operator can dramatically alter the cost required by tabu search to locate optimal solutions to random JSPs. In many cases, the effect is actually *detrimental* in that the mean number of iterations required under $TS_{N5}$ can be significantly larger than that required under $TS_{N1}$. However, the number of neighbors under the $N1$ operator commonly exceeds that under the $N5$ operator by a factor of 10 or more, especially on larger problem instances, masking any detrimental effects in the vast majority of cases. As a result, $TS_{N5}$ consistently locates optimal solutions in lower overall *run-times* on average than $TS_{N1}$.

Large differences in the observed $\bar{c}$ under $TS_{N1}$ and $TS_{N5}$ are necessarily indicative of differences in the underlying run-time dynamics. We now consider whether the differences are truly qualitative or merely quantitative. In other words, is the random walk model proposed in Section 7 no longer applicable, or can the differences be explained in terms of changes in model parameters such as $D_{max}$ and the transition probabilities $P(S_{j,x'}|S_{i,x})$? To answer this question, we first compute estimates of the dynamic cost model parameters using the sampling methodology described in Section 7.2, with one exception: due to their relative rarity, we do not attempt to capture the random walk events associated with trap escape. To ensure tractability, we consider only the 42 of our 10 × 10 random JSPs with ≤ 100,000 optimal solutions.

The resulting transition probabilities are more irregular than those observed under $TS_{N1}$ on the same problem instances, mirroring the results obtained for structured JSPs described in Section 10. Additionally, we frequently observe large discrepancies in $D_{max}$ under $TS_{N1}$ and $TS_{N5}$, which, in part, accounts for the observed discrepancies in $\overline{c}$. Using the resulting parameter estimates, we compute the predicted $\overline{c}$ and compare the results with the actual $\overline{c}$ observed using $TS_{N5}$; the corresponding scatter-plot is shown in the right side of Figure 11. The $r^2$ value of the corresponding $log_{10}$-$log_{10}$ regression model is 0.96, and in all cases the actual $\overline{c}$ deviates from the predicted $\overline{c}$ by no more than a factor of 5. Overall, these results demonstrate that the N5 move operator has negligible effect on the absolute accuracy of the dynamic cost model; as in $TS_{N1}$, search in $TS_{N5}$ simply acts as a biased random walk over the $S_{lopt+}$ sub-space. Consequently, the dynamic cost model is an appropriate basis for a detailed analysis of $i$-TSAB or related algorithms, which differ from $TS_{N5}$ primarily in the use of long-term memory mechanisms such as reintensification and path relinking (Nowicki & Smutnicki, 2005).

## 12. Exploring the Predictive Capability of the Dynamic Cost Model

Thus far, our primary goal has been to explain the source of the variability in the cost of locating optimal solutions to random JSPs using $TS_{N1}$; the dynamic cost model introduced in Section 7 largely achieves this objective. Despite this success, however, we have only illustrated the explanatory power of the model. Ideally, scientific models are predictive, in that they lead to new conjectures concerning subject behavior and are consequently falsifiable. We next use the dynamic cost model to propose and empirically confirm two novel conjectures regarding the behavior of $TS_{N1}$ on random JSPs. Our analysis demonstrates that the utility of cost models can extend beyond after-the-fact explanations of algorithm behavior.

### 12.1 The Variable Benefit of Alternative Initialization Methods

Empirical evidence suggests that high-quality initial solutions *can* improve the performance of tabu search algorithms for the JSP (Jain, Rangaswamy, & Meeran, 2000). Yet, both the exact conditions under which improvements can be achieved and the expected degree of improvement are poorly understood. We now explore a particular aspect of this broader issue by considering the question: What impact do different initialization methods have on the cost required by $TS_{N1}$ to locate optimal solutions to random JSPs? The preceding analyses of $TS_{N1}$ are based on the assumption that search is initiated from a random local optimum. Here, we instead consider the behavior predicted by the dynamic cost model when search is initiated from solutions other than random local optima.

Demystifying Tabu Search

Let $\overline{v}_i$ denote the predicted mean search cost required to locate an optimal solution under the dynamic cost model when search is initiated from solutions that are distance $i$ from the nearest optimal solution. As in Section 7, we assume the initial gradient $x$ equals *closer* or *farther* with equal probability. Ignoring potential asymmetries in the distribution of $d_{opt}(s)$ for random local optima $s$, observe that the predicted $\overline{c}$ is approximately equal to $\overline{v}_\delta$ where $\delta = rint(\overline{d}_{lopt\text{-}opt})$, i.e., $TS_{N1}$ is initiated from a local optimum that is an average distance $\overline{d}_{lopt\text{-}opt}$ from the nearest optimal solution. We address the issue of the impact of alternative initial solutions on the performance of $TS_{N1}$ by analyzing the nature of $\overline{v}_i$ for $i \neq \delta$. In Figure 12, we show plots of the predicted costs $\overline{v}_i$ over a wide range of $i$ for specific $6 \times 6$ (left figure) and $10 \times 10$ (right figure) random JSPs; results for $i \leq 2$, where $\overline{v}_i \ll 100$, are omitted for purposes of visualization. For the $6 \times 6$ instance, search cost rises rapidly between $i = 3$ and $i = 10$ and continues to gradually increase as $i \to D_{max}$. In contrast, search cost for the $10 \times 10$ instance rises rapidly between $i = 2$ and $i \approx 15$ but is roughly *constant*, modulo the sampling noise, once $i > 15$. Even when $i = 3$, the dynamic model predicts that search cost is still significant: if the initial search gradient is not *closer*, search is rapidly driven toward solutions that are distant from the nearest optimal solution and any benefit of the favorable initial position is lost. We observe qualitatively identical behavior in a large sample of our random JSPs, arriving at the following general observation: for easy (hard) instances, the approach toward an asymptotic value of $v_i$ as $i \to D_{max}$ is gradual (rapid). Consequently, we hypothesize that a particular initialization method will at best have a minimal impact on the performance of $TS_{N1}$ *unless* the resulting solutions are very close to the nearest optimal solution. Additionally, we observe that the dynamic cost model predicts that the distance to the nearest optimal solution, and not solution fitness, dictates the benefit of a given initialization method. The distinction is especially key in the JSP, where fitness-distance correlation is known to be comparatively weak, e.g., in contrast to the Traveling Salesman Problem (Mattfeld, 1996).

To test this hypothesis, we analyze the performance of $TS_{N1}$ using a variety of heuristic and random methods to generate initial solutions. Following Jain et al. (2000), we consider the following set of high-quality priority dispatch rules (PDRs) used in conjunction with Giffler and Thompson's (1960) procedure for generating active solutions[8]:

- *FCFS* (First-Come, First-Serve),
- *LRM* (Longest ReMaining work),
- *MWKR* (Most WorK Remaining), and
- *SPT* (Shortest Processing Time).

Additionally, we consider both active and non-delay solutions (Giffler & Thompson, 1960) generated using random PDRs, respectively denoted $RND_{actv}$ and $RND_{ndly}$. Finally, we include Taillard's (1994) lexicographic construction method, denoted *LEX*, and the insertion procedure introduced by Werner and Winkler (1995), which we denote *WW*; the latter is one of the best constructive heuristics available for the random JSP (Jain et al., 2000). Random semi-active solutions serve as a baseline and are denoted $RND_{semi}$. The solutions resulting from all methods are transformed into local optima by applying steepest-descent under the *N1* operator.

---

8. Technically, there is a formal difference between a solution and a schedule in the JSP. However, because we are assuming earliest start-time scheduling of all operations, we use the two terms interchangeably.

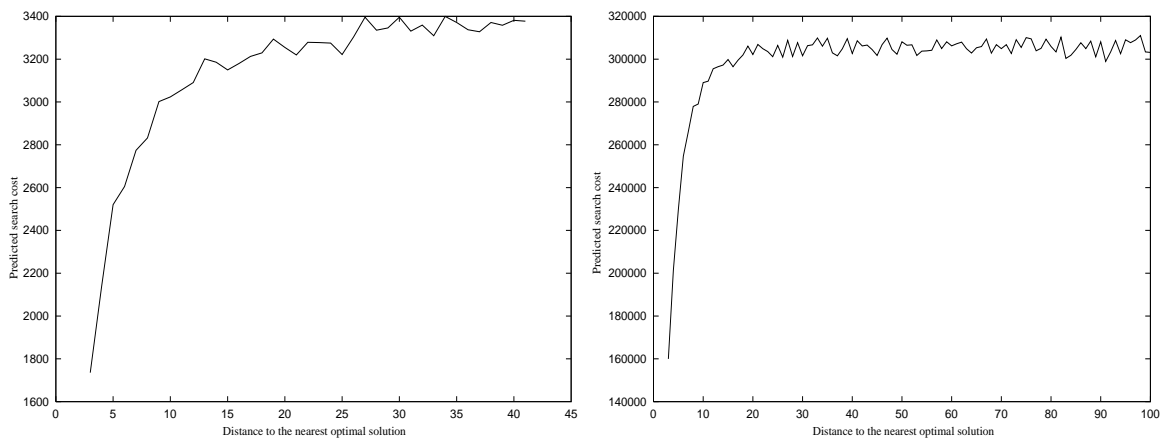

Figure 12: Plots of the distance $i$ between an initial solution and the nearest optimal solution and the predicted search cost $\overline{v}_i$ for a $6 \times 6$ (left figure) and a $10 \times 10$ (right figure) random JSP. The $\delta = rint(\overline{d}_{lopt\text{-}opt})$ for these two instances are 29 and 65, respectively.

We again consider only the 42 $10 \times 10$ random JSPs with $\leq$ 100,000 optimal solutions; we selected the larger - as opposed to $6 \times 4$ or $6 \times 6$ - problem set due to the higher degree of expected difficulty. For each initialization method, we compute $\overline{d}_{lopt\text{-}opt}$ for each problem instance using (with the exception of *LEX*, which is deterministic) 5,000 local optima generated by applying steepest-descent search to the solutions resulting from the given method. For $RND_{semi}$, we obtain a mean $\overline{d}_{lopt\text{-}opt}$, averaged over all 42 instances, of approximately 70.92. We show the mean $\overline{d}_{lopt\text{-}opt}$ for each remaining initialization method in Table 4; p-values for the statistical significance of the mean difference in $\overline{d}_{lopt\text{-}opt}$ between the various methods and $RND_{semi}$, computed using a Wilcoxon non-parametric, paired-sample signed rank test, are also provided. With the exception of *SPT*, we observe significant differences in $\overline{d}_{lopt\text{-}opt}$ between the baseline $RND_{semi}$ solutions and those resulting from other initialization methods. Initially, these data suggest that it may be possible to improve the performance of $TS_{N1}$ using initialization methods with low $\overline{d}_{lopt\text{-}opt}$. However, the lowest mean values of $\overline{d}_{lopt\text{-}opt}$, obtained using the *LEX* and *WW* methods, are still large in absolute terms. Consequently, given a combination of our working hypothesis and the average difficulty of $10 \times 10$ instances (e.g., see the right side of Figure 12), it seems likely that even these solutions are still too far from the nearest optimal solution to impact overall search cost.

For each problem instance, we next compute the $c_{Q2}$ under each initialization method; statistics are taken over 1,000 independent trials of $TS_{N1}$. The percent differences in $c_{Q2}$ for each initialization method relative to that obtained under the $RND_{semi}$ baseline are reported in Table 4. The *worst-case* deviation is less than 3% and the best improvement, obtained under *WW*, is only 2.79%. Further, all observed discrepancies can be attributed to sampling error in computation of $c_{Q2}$ and no differences were statistically significant even at $p \leq 0.05$. The data support the hypothesis predicted by the dynamic cost model: for difficult problems, available initialization methods for the JSP have *no* significant impact on the performance of $TS_{N1}$. We conclude by observing that our results say nothing about the cost required for $TS_{N1}$ to locate optimal solutions to easy-to-moderate instances or sub-optimal solutions on a range of problem instances. In particular, we observe that alternative initialization methods may improve performance in these situations, due to the gradual increase of $\overline{v}_i$ associated with less difficult problem instances. Similarly, alternative initialization methods may benefit tabu search algorithms that employ re-intensification, such as those developed by Nowicki and Smutnicki. We have not investigated whether similar results hold on structured JSPs, principally because of the increased difficulty in computing both $c_{Q2}$ and $\overline{d}_{lopt\text{-}opt}$ for these instances.

| | Initialization Method | | | | | | | |
|---|---|---|---|---|---|---|---|---|
| | $FCFS$ | $LRM$ | $MWKR$ | $SPT$ | $LEX$ | $RND_{actv}$ | $RND_{ndly}$ | $WW$ |
| $\overline{d}_{lopt\text{-}opt}$ | 58.49 | 97.41 | 97.94 | 64.97 | 49.25 | 64.68 | 58.55 | 53.10 |
| $p$ for mean difference in $\overline{d}_{lopt\text{-}opt}$ relative to $RND_{semi}$ | 0.001 | 0.001 | 0.001 | 0.126 | 0.001 | 0.001 | 0.001 | 0.001 |
| % mean difference in $c_{Q2}$ relative to $RND_{semi}$ | 1.76 | 2.32 | 2.94 | 1.55 | 1.44 | 1.07 | 0.06 | -2.79 |
| $p$ of mean difference in $log_{10}(c_{Q2})$ relative to $RND_{semi}$ | 0.059 | 0.084 | 0.073 | 0.077 | 0.513 | 0.573 | 0.556 | 0.309 |

Table 4: The differences in both the mean distance to the nearest optimal solution ($\overline{d}_{lopt\text{-}opt}$) and search cost ($c_{Q2}$) for various initialization methods on $10 \times 10$ random JSPs; differences are measured relative to random semi-active solutions ($RND_{semi}$).

## 12.2 The Specification of Tabu Tenure

Empirically, the performance of tabu search depends heavily upon the choice of tabu tenure. Although "no single rule has been designed to yield an effective tenure for all classes of problem" (Glover & Laguna, 1997, p. 47), it is generally recognized that small tabu tenures lead to search stagnation, i.e., the inability to escape local optima, while large tabu tenures can yield significant deterioration in search effectiveness. Beyond these loose observations, practitioners have little guidance in selecting tabu tenures, and there is no theoretical justification for preferring any particular values within a range of apparently reasonable possibilities. In $TS_{N1}$, a side-effect of short-term memory is to consistently bias search either toward or away from the nearest optimal solution. Intuitively, we would expect the magnitude of this bias to be proportional to the tabu tenure $L$; longer tenures should force search to make more rapid progress away from previously visited regions of the search space. Consider the scenario in which $TS_{N1}$ is steadily progressing away from the nearest optimal solution, such that the current distance to the nearest optimal solution is given by $X \gg \overline{d}_{tabu\text{-}opt}$. For any fixed $L$, $TS_{N1}$ will eventually invert the gradient and move search toward the nearest optimal solution. However, the larger the value of $L$, the more distant $TS_{N1}$ is likely to move from the nearest optimal solution before inversion occurs. In terms of our dynamic model of $TS_{N1}$, this suggests that the maximal likely distance $D_{max}$ to the nearest optimal solution achieved by $TS_{N1}$ is proportional to $L$. We have shown that problem difficulty in the JSP is largely a function of the effective search space size $|S_{lopt+}|'$, of which $D_{max}$ is one measure. Consequently, we hypothesize that any increase in the tabu tenure translates into growth in $|S_{lopt+}|'$ and by inference problem difficulty.

To test this hypothesis, we first examine the $D_{max}$ obtained using our sampling methodology over a range of tabu tenures. We consider $10 \times 10$ random JSPs, specifically the 42 instances with $\leq 100,000$ optimal solutions. In $TS_{N1}$, the mean tabu tenure $L$ is dictated by the interval $[L_{min}, L_{max}]$. Previously, we let $[L_{min}, L_{max}] = [8, 14]$; this particular value was empirically determined by Taillard to yield good performance on the `ft10` $10 \times 10$ benchmark instance. We now test the impact of both smaller and larger tabu tenure intervals on the performance of $TS_{N1}$. Based on extensive experimentation, we observe that $[5, 10]$ approximates the smallest tenure interval for which $TS_{N1}$ is empirically PAC, i.e., avoids becoming trapped in either local optima or restricted regions of the search space. On average, the $D_{max}$ obtained under the $[8, 14]$ interval are roughly 6% greater than those obtained under the $[5, 10]$ interval, while the $D_{max}$ obtained under a larger (arbitrarily chosen) $[10, 18]$ interval are in turn roughly 5% greater than those observed under the $[8, 14]$ interval; we attribute the non-uniform growth rate to differences in the mean tabu tenures under the $[5, 10]$ and $[8, 14]$ intervals versus the $[8, 14]$ and $[10, 16]$ intervals. Overall, these results confirm the intuition that larger tabu tenures lead to increased $|S_{lopt+}|'$, as measured by $D_{max}$; we observe qualitatively identical changes in $\overline{d}_{tabu\text{-}opt}$.

To confirm that changes in $D_{max}$ yield corresponding changes in problem difficulty, we compute the observed $\overline{c}$ under $TS_{N1}$ for each problem instance over the three tabu tenure intervals. Here, we consider all 100 instances in our $10 \times 10$ problem set; the implicit assumption is that similar changes in $D_{max}$ hold for instances with $> 100,000$ optimal solutions. Scatter-plots of the resulting $\overline{c}$ for $[5, 10]$ versus $[8, 14]$ and $[8, 14]$ versus $[10, 18]$ tenure intervals are respectively shown in the left and right sides of Figure 13. The $\overline{c}$ under the medium interval $[8, 14]$ are roughly 95% larger than those observed under the smaller $[5, 10]$ interval, while the $\overline{c}$ obtained under the $[10, 16]$ interval are roughly 60% greater than those obtained under the $[8, 14]$ interval; again, we attribute the non-uniform growth rate to discrepancies in the difference in mean tabu tenure. We observe similar monotonic growth in problem difficulty for a limited sample of even larger tenure intervals. Overall, the results support our hypothesis that larger tabu tenures increase problem difficulty, specifically by inflating $|S_{lopt+}|'$. Although not reported here, we additionally observe similar results on a small sample of $6 \times 6$ workflow and flowshop JSPs.

Our experiments indicate that the tabu tenure $L$ for $TS_{N1}$ should be chosen *as small as possible* while simultaneously avoiding search stagnation. In addition to providing the first theoretically justified guideline for selecting a tabu tenure, this observation emphasizes the potentially detrimental nature of short-term memory. In particular, the results presented above suggest that *any* amount of short-term memory in excess of that which is required to escape the attractor basins of local optima is likely to degrade the performance of $TS_{N1}$.

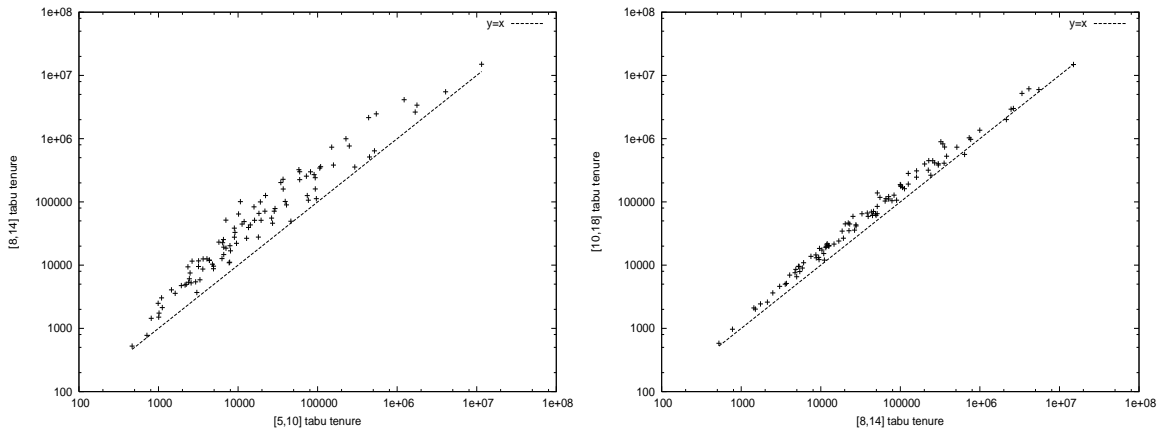

Figure 13: Scatter-plots of the relative cost $\overline{c}$ required to locate an optimal solution under $TS_{N1}$ for small versus moderate tabu tenures (left figure) and moderate versus large tabu tenures (right figure), for $10 \times 10$ random JSPs.

## 13. Implications and Conclusions

Our results provide a significant first step toward developing an understanding of the dynamics underlying tabu search. We have introduced a random walk model of Taillard's tabu search algorithm for the JSP that, despite its simplicity, accounts for nearly all of the variability in the cost required to locate optimal solutions to random JSPs. Additionally, the model accounts for similarly high proportions of the variability in the cost required to locate both sub-optimal solutions to random JSPs and optimal solutions to more structured JSPs. Our results indicate that search in Taillard's algorithm can be viewed as a variant of a straightforward one-dimensional random walk that exhibits two key types of bias: (1) a bias toward solutions that are roughly equi-distant from the nearest optimal solution and solutions that are maximally distant from the nearest optimal solution and (2) a bias that tends to maintain consistent progress either toward or away from the nearest optimal solution. In contrast to cost models of problem difficulty based on static search space features, the random walk model is scalable and provides direct insight into the dynamics of the search process. Additionally, we identified an unexpectedly strong link between the run-time dynamics of tabu search and simple features of the underlying search space, which provides an explanation for the initial success and ultimate failure of our earlier $\overline{d}_{lopt\text{-}opt}$ model of problem difficulty. Although we have not fully explored the predictive capabilities of the random walk model, two novel behaviors predicted by the model have been confirmed through experimentation on random JSPs: the failure of initial solutions to significantly impact algorithm performance and the potentially detrimental nature of short-term memory.

Despite the success of the random walk model, several issues remain unresolved. For example, it is unclear why $TS_{N1}$ is unlikely to explore potentially large regions of $S_{lopt+}$, i.e., why random local optima are not necessarily representative of solutions visited by $TS_{N1}$ during search. Similarly, a causal explanation for the bias toward solutions that are roughly equi-distant from optimal solutions and solutions maximally distant from optimal solutions is lacking, although preliminary evidence indicates the bias is simply due to the choice of representation, i.e., the binary hypercube.

Perhaps the most important contribution of the random walk model is the foundation it provides for future research. State-of-the-art tabu search algorithms for the JSP make extensive use of long-term memory (Nowicki & Smutnicki, 2005), and it is unclear how such memory will impact the structure of the random walk model. Moving beyond the JSP, there is the question of generalization: Do similar results hold when considering tabu search algorithms for other $NP$-hard problems, e.g., the quadratic assignment and permutation flow-shop scheduling problems? Although the model and associated methodology can be straightforwardly applied to other problems, representations, and even local search algorithms, it is unclear *a priori* whether we can expect sufficient regularities in the resulting transition probabilities to yield accurate predictions. This is especially true for problems in which highly structured benchmarks are more prevalent, e.g., in SAT. Finally, the random walk model is, at least currently, largely of only *a posteriori* use. It is unclear how such a model might be leveraged in order to develop improved tabu search algorithms. For example, although it is clear that the bias toward solutions that are distant from optimal solutions should be minimized, it is far from obvious how this can be achieved. Similarly, another potential application of the random walk model involves predicting problem difficulty; now that the dominant factors influencing problem difficulty in the JSP are becoming better understood, an obvious next step is to analyze whether it is possible to achieve accurate estimates of these quantities with minimal or moderate computational effort.

The objective of our research was to "demystify" the behavior of tabu search algorithms, using the JSP as a testbed. In this goal, we have succeeded. Our random walk model captures the run-time dynamics of tabu search for the JSP accounts for the primary behavior of interest: the cost required to locate optimal solutions to problem instances. The power of the model is further illustrated by its ability to account for additional behavioral phenomena and correctly predict novel behaviors. Through careful modeling and analysis we have demonstrated that despite their effectiveness, tabu search algorithms for the JSP are in fact quite simple in their operation. The random walk model should serve as a useful basis for exploring similar issues in the context of both more advanced tabu search algorithms for the JSP and tabu search algorithms for other $NP$-hard combinatorial optimization problems.

## Acknowledgments

Sandia is a multipurpose laboratory operated by Sandia Corporation, a Lockheed-Martin Company, for the United States Department of Energy under contract DE-AC04-94AL85000. This work was sponsored in part by the Air Force Office of Scientific Research, Air Force Materiel Command, USAF, under grant number F49620-03-1-0233. The U.S. Government is authorized to reproduce and distribute reprints for Governmental purposes notwithstand-ing any copyright notation thereon. The authors would like to thank Matthew Streeter for pointing out a flaw with an earlier version of our methodology for estimating the transition probabilities in the dynamic cost model. Finally, we gratefully acknowledge the exceptionally thoughtful and detailed feedback from the anonymous reviewers, which enabled us to significantly improve our original presentation.